%% file: acl_latex.tex
\documentclass[11pt, a4paper, copyright, gdm]{google}

\usepackage[authoryear, sort&compress, round]{natbib}
\uselogo{}

\usepackage{amsmath}
\usepackage{amssymb}
\usepackage{bm}

\input{math_commands.tex}

\usepackage{times}
\usepackage{latexsym}
\usepackage{hyperref}
\usepackage{booktabs}
\usepackage{xspace}
\usepackage[table]{xcolor}

\let\cite\citep

\usepackage[T1]{fontenc}

\usepackage[utf8]{inputenc}

\usepackage{microtype}

\usepackage{inconsolata}

\usepackage{hyperref}
\usepackage{url}
\usepackage{xurl} 
\usepackage{booktabs,tabularx}
\usepackage{graphicx}
\usepackage{booktabs}
\usepackage{multirow}
\usepackage{arydshln}
\usepackage[most]{tcolorbox}
\usepackage{xcolor}
\usepackage{enumitem}
\usepackage{subcaption}
\usepackage[font=small,labelfont=bf]{caption}
\definecolor{countryColor}{RGB}{0,153,51}   
\definecolor{ruleColor}{RGB}{0,102,204}     
\definecolor{exampleColor}{RGB}{204,0,102}  
\definecolor{ctxColor}{RGB}{0,102,204}       
\definecolor{varColor}{RGB}{0,153,51}        
\definecolor{warnColor}{RGB}{204,0,102}      
\definecolor{boxBg}{RGB}{248,248,248}        
\tcbset{sharp corners, boxrule=0.5pt}
\definecolor{facetColor}{HTML}{D35400}

\title{Steering LLMs for Culturally Localized Generation}


\author{
Simran Khanuja$^{*2}$, Hongbin Liu$^{1}$, Shujian Zhang$^{1}$, John Lambert$^{1}$, Mingqing Chen$^{1}$, Rajiv Mathews$^{1}$, Lun Wang$^{1}$ \\
$^{1}$Google DeepMind, $^{2}$Carnegie Mellon University, $^{*}$Work done as a student researcher at Google DeepMind. \\
Correspondence to \texttt{skhanuja@andrew.cmu.edu}
}

\begin{document}

\input{defs}
\input{sections/0-abstract}

\maketitle

\input{sections/1-introduction}

\input{sections/3-methodology}

\input{sections/4-experiments}
\input{sections/5-results}
\input{sections/2-related_work}

\input{sections/6-conclusion}

\input{sections/7-ethics_limitations}

\bibliography{custom}

\newpage
\appendix
\label{sec:app}
\section{Why augment the CANDLE dataset?}
\label{sec:augment}
Upon manually inspecting the CANDLE dataset, we found artifacts that could possibly affect the quality of features selected. Most notably, we saw that many CANDLE assertions are common and popular facts (like \emph{The maple leaf is on the Canada flag}) and each assertion has the country name in it or a variant (like American, USA, US, United States etc.) given how they collect their data \cite{nguyen2023extracting} (\emph{We start the extraction by searching for sentences that contain mentions of the given subjects}). The CANDLE dataset is still valuable to include since it is well-categorized naturally-occurring data. We prompt a LLM to augment this data such that it does not contain the artifacts in CANDLE (like country names, very generic facts etc.) to increase data diversity and improve feature discovery. We also provide the chosen CANDLE examples in-context to the model during generation to avoid duplication. The exact prompt we use can be found in \ref{fig:candle_augmentation_prompt}.

We also ran an ablation to understand which features are selected when using the CANDLE dataset alone.  Following the procedure described in Section \ref{sec:cue}, we select features whose cumulative MI reaches a threshold of 0.1.

Using only CANDLE results in selecting 4,473 features (37.4\% of the total feature set of 11,961), with 888 features shared with the full dataset. This analysis reveals three key observations: \textbf{i)} \emph{Top mutual-information features are stable} -- The highest-MI features are largely consistent across both datasets, while lower-ranked features vary depending on the dataset used; \textbf{ii)} \emph{Marginal features are dataset-dependent} -- Many features selected only in the CANDLE-only setting appear to reflect surface artifacts in the data; \textbf{iii)} \emph{Country-name artifacts activate shallow layers} -- The largest difference occurs in the earliest layer. CANDLE alone selects 14.11\% of features from layer 0, compared to only 0.17\% when using the augmented dataset. Because each CANDLE assertion explicitly contains the country name, this likely activates shallow lexical features encoding the country identifier rather than deeper cultural or semantic representations. The augmented data reduces this effect by removing explicit country references.

Table~\ref{tab:candle_layer_distribution} shows the layer-wise distribution of selected features.

\begin{table}[h]
\centering
\small
\begin{tabular}{lcc}
\toprule
\textbf{Layer} & \textbf{CANDLE Only} & \textbf{Augmented Dataset} \\
\midrule
0  & 14.11\% & 0.17\% \\
1  & 4.20\%  & 0.44\% \\
2  & 5.63\%  & 1.02\% \\
3  & 3.13\%  & 0.78\% \\
4  & 5.37\%  & 1.28\% \\
5  & 3.80\%  & 2.01\% \\
6--12  & $\sim$3.9--6.5\% & $\sim$3.3--4.1\% \\
13--15 & $\sim$3.0--3.4\% & $\sim$4.7--4.9\% \\
16--24 & 1.7--2.9\% & 5.0--7.1\% \\
25 & 2.86\% & 3.17\% \\
\bottomrule
\end{tabular}

\caption{Layer-wise distribution of selected features when using the CANDLE dataset alone versus the augmented dataset for Gemma-2-2B. The CANDLE-only setting disproportionately activates early-layer features due to explicit country-name tokens present in every assertion.}
\label{tab:candle_layer_distribution}

\end{table}

In Section \ref{sec:step0}, we mention that we rewrite prompts before obtaining classification results so that the model does not rely on trivial cues like country names to make its prediction. The prompt to rewrite assertions is given in Figure \ref{fig:country_neutral_prompt}.

\begin{figure*}[t]
\centering
\small
\setlist[itemize]{noitemsep, topsep=2pt, leftmargin=1.3em}

\begin{minipage}{0.95\linewidth}

\textbf{Prompt Template: CANDLE Augmentation with Uncommon Assertions}

\medskip
\textbf{Role.}  
You are an expert on world cultures. Generate
\textcolor{varColor}{\texttt{\{num\_assertions\}}} factual assertions about
\textcolor{countryColor}{\texttt{\{country\}}} that are culturally specific yet not easily identifiable from explicit country markers.

\medskip
\textbf{Task.}  
Produce exactly
\textcolor{varColor}{\texttt{\{num\_assertions\}}} new, unique, and factually accurate assertions about
\textcolor{countryColor}{\texttt{\{country\}}}, optionally focusing on the facet
\textcolor{facetColor}{\texttt{\{facet\}}}.  
Each assertion must be a complete, self-contained sentence.

\medskip
\textbf{Requirements.}

\begin{itemize}
\item \textbf{Uncommon / lesser-known facts only.} Avoid obvious clichés and widely known facts. Instead focus on:
\begin{itemize}
\item Specific customs that are not internationally famous
\item Subtle social norms and unwritten rules
\item Regional cultural variations
\item Historical traditions still practiced today
\item Unique aspects of everyday life
\item Lesser-known food or drink customs
\item Specific etiquette or behavioral norms
\end{itemize}

\item \textbf{No country identifiers.} Assertions must \textbf{not} contain:
\begin{itemize}
\item The country name (\texttt{\{country\}})
\item Nationality adjectives (e.g., American, Chinese, French)
\item City or region names within the country
\item Highly identifiable landmarks or institutions
\end{itemize}

\item \textbf{Factually accurate.} Each statement should be verifiable.

\item \textbf{Self-contained.} Each assertion should stand alone as a full sentence.

\item Avoid generating assertions similar to existing dataset examples.
\end{itemize}

\medskip
\textbf{Facet Instruction.}

\quad \texttt{\{facet\_instruction\}}

\smallskip
\textbf{Facet Options.}  
food, drink, traditions, customs, daily life, values, social norms, celebrations.

\medskip
\textbf{Examples of Good Assertions (Style Reference Only).}

\begin{itemize}
\item ``It is considered rude to tip at restaurants, as service is included in the price.''
\item ``Slurping noodles loudly is a sign of appreciation for the meal.''
\item ``Removing shoes before entering a home is expected, and hosts often provide slippers.''
\item ``The number 4 is avoided in elevators and floor numbering due to its association with death.''
\item ``Tea is traditionally served in small glasses rather than cups.''
\end{itemize}

\medskip
\textbf{Bad Examples.}

\begin{itemize}
\item ``This country is famous for its cuisine.'' (too vague)
\item ``French wine is world-renowned.'' (contains nationality)
\item ``People in Tokyo take the train to work.'' (contains city name)
\item ``The Great Wall is a famous landmark.'' (too obvious)
\end{itemize}

\medskip
\textbf{Avoid generating assertions similar to these existing examples.}

\quad \texttt{\{existing\_examples\}}

\medskip
\textbf{Output Format.}

\begin{itemize}
\item Generate exactly \texttt{\{num\_assertions\}} assertions
\item Output one assertion per line
\item Do not number the assertions
\item Do not include extra commentary or explanation
\end{itemize}

\end{minipage}

\caption{Prompt template used to augment the CANDLE dataset with culturally specific but non-identifying assertions.}
\label{fig:candle_augmentation_prompt}

\end{figure*}






\section{Toy example illustrating \method}
\label{app:toy_example}
Our dataset consists of $(\text{assertion}, \text{culture label})$ pairs.
We compute mutual information (MI) globally between each SAE feature
activation and the culture label. The selected features are those whose
activations reduce uncertainty about the country label. All countries are
represented using the same selected feature set $\mathcal{S}$, while
activation magnitudes differ across countries. Note that the following example is purely illustrative used only to demonstrate the mechanics of the method. They do not correspond to
actual SAE features, dataset entries, or empirical measurements in our
experiments.

Suppose a sparse autoencoder discovers four interpretable features:

\begin{center}
\begin{tabular}{ll}
\toprule
Feature & Interpreted meaning \\
\midrule
$F_1$ & rice-related \\
$F_2$ & baseball-related \\
$F_3$ & tea-related \\
$F_4$ & breakfast-related \\
\bottomrule
\end{tabular}
\end{center}

Assume three countries with three assertions each.
\begin{table}[h]
\centering
\small
\begin{tabularx}{\columnwidth}{l X l}
\toprule
Country & Assertion & Act. fts. \\
\midrule
Japan & Rice is eaten daily & $F_1$ \\
Japan & Green tea and rice are common & $F_1, F_3$ \\
Japan & People eat breakfast in the morning & $F_4$ \\
\midrule
US & Baseball games serve hot dogs & $F_2$ \\
US & Baseball is very popular & $F_2$ \\
US & The first meal of the day is breakfast & $F_4$ \\
\midrule
UK & Tea is common at breakfast & $F_3$ \\
UK & Tea is served in the afternoon & $F_3$ \\
UK & People wake up and eat breakfast & $F_4$ \\
\bottomrule
\end{tabularx}
\caption{Hypothetical dataset illustrating feature activations in the toy example}
\label{tab:toy_dataset}
\end{table}

Note that the breakfast assertion appears across all countries. This
activates a feature common to every class.

\paragraph{Step 1: Feature activations} -- After extracting SAE activations and applying max pooling, we obtain:

\begin{table}[h]
\centering
\small
\begin{tabular}{lccccl}
\toprule
Assertion & $F_1$ & $F_2$ & $F_3$ & $F_4$ & Country \\
\midrule
Rice eaten daily & 8 & 0 & 0 & 0 & Japan \\
Green tea \& rice & 6 & 0 & 5 & 0 & Japan \\
Breakfast statement & 0 & 0 & 0 & 7 & Japan \\
Baseball hot dogs & 0 & 7 & 0 & 0 & US \\
Baseball popular & 0 & 6 & 0 & 0 & US \\
Breakfast statement & 0 & 0 & 0 & 7 & US \\
Tea breakfast & 0 & 0 & 8 & 0 & UK \\
Tea afternoon & 0 & 0 & 7 & 0 & UK \\
Breakfast statement & 0 & 0 & 0 & 7 & UK \\
\bottomrule
\end{tabular}
\caption{Toy SAE activations.}
\end{table}

We compute mutual information between each feature and the country label.
Intuitively:

\begin{itemize}
\item $F_1$ (rice) activates primarily for Japan $\rightarrow$ informative
\item $F_2$ (baseball) activates only for the US $\rightarrow$ informative
\item $F_3$ (tea) activates for Japan and the UK $\rightarrow$ partially informative
\item $F_4$ (breakfast) activates equally across all countries $\rightarrow$ uninformative
\end{itemize}

Since $F_4$ does not reduce uncertainty about the country label, its MI is
approximately zero and it is not selected. Thus the globally selected
feature set is

\[
\mathcal{S} = \{F_1, F_2, F_3\}.
\]

\paragraph{Step 2: Construct country prototypes} -- For each country $c \in \mathcal{C}$ we compute the prototype

\[
\mathbf{p}_{\method}^{(c)} =
\frac{1}{N_c} \sum_{x \in \mathcal{D}_c} \method(x).
\]

Restricting activations to $\mathcal{S}$ yields the prototype matrix

\[
P_{\method} =
\begin{bmatrix}
4.67 & 0 & 1.67 \\
0 & 4.33 & 0 \\
0 & 0 & 5.00
\end{bmatrix}
\]

where rows correspond to Japan, the United States, and the United Kingdom
respectively.

\paragraph{Step 3a: Bias analysis} -- To isolate relative cultural signals, we subtract the mean prototype across countries:

\[
\boldsymbol{\mu}_{\text{global}} = [1.56, 1.44, 2.22].
\]

We center each prototype

\[
\tilde{P} =
\begin{bmatrix}
3.11 & -1.44 & -0.55 \\
-1.56 & 2.89 & -2.22 \\
-1.56 & -1.44 & 2.78
\end{bmatrix}.
\]

Suppose the model generates the response -- \emph{A warm cup of tea.} Its SAE activation is $a(y) = [0,0,8]$. After centering,

\[
\tilde{a}(y) =
a(y)[\mathcal{S}] - \boldsymbol{\mu}_{\text{global}}
=
[-1.56, -1.44, 5.78].
\]

We compute cultural alignment using cosine similarity: $\text{bias}(y,c) =\cos(\tilde{a}(y), \tilde{p}^{(c)})$.

\[
\text{bias}(y,\text{Japan}) = -0.279
\]
\[
\text{bias}(y,\text{US}) = -0.598
\]
\[
\text{bias}(y,\text{UK}) = 0.955
\]

Thus the generated response aligns most strongly with the United Kingdom
prototype.

\section{Hyperparameter Selection}
\label{app:hparam}

All experiments are conducted using pretrained language models and pretrained sparse autoencoders (SAEs). We use publicly available SAE checkpoints from the Gemma Scope \cite{lieberum2024gemma} and Llama Scope \cite{he2024llama} releases, which provide SAEs trained across all layers of chosen models for certain SAE widths. This allows us to analyze the layer-wise distribution of culturally informative features identified by \method.

For \texttt{Gemma-2-2B-16K}, \texttt{Gemma-2-9B-16K}, and \texttt{LLaMA-3.1-8B-32K}, we load SAEs for all layers of the model and apply steering at every layer. This design choice enables us to later inspect how many features the algorithm selects from each layer. For larger SAE widths and larger models (\texttt{Gemma-2-9B-131K}, and \texttt{LLaMA-3.1-8B-128K}), loading SAEs
for all layers simultaneously exceeds the memory capacity of a single NVIDIA A100 (80GB). In these cases we load SAEs at every fourth layer (i.e., layers $0,4,8,12,\ldots$) and apply steering only at those layers.

We select culturally informative SAE features using a cumulative mutual information threshold $\rho$. Unless otherwise stated, we fix $\rho = 0.1$, corresponding to selecting the top 10\% of features ranked by mutual information across all layers. The steering coefficient $\alpha$ controls the magnitude of the intervention in the residual stream. We perform a sweep over
$\alpha \in \{0.25, 0.5, 1, 2\}$ for each target culture. The best value of $\alpha$ is selected independently for each country, as different countries exhibit different prototype magnitudes and feature activation densities.

All generations are produced with temperature $0.9$ unless otherwise
stated. All experiments are conducted on a single NVIDIA A100 GPU with 80GB of
memory.

\section{Evaluation Prompts}
\label{app:eval_prompts}

This section provides the exact prompts used for LLM-based evaluation.
As described in Section~\ref{sec:setup}, responses are evaluated using an
LLM-as-a-judge framework with two independent judges. The prompts below
are used to score generations along the dimensions reported in the
paper.

Figure~\ref{fig:prompt_eval_all} shows the prompt used for individual
evaluation of generated responses, where judges assign Likert-scale
scores for cultural faithfulness, fluency, and rarity.  
Figure~\ref{fig:prompt_pairwise_eval} shows the pairwise evaluation
prompt used to compare competing generations. In pairwise comparisons,
judges score both responses independently and the winner is determined
based on these scores.

The prompts are shown verbatim to ensure full reproducibility of the
evaluation procedure.

\begin{figure*}[t]
\centering
\begin{tcolorbox}[colframe=black!70, colback=boxBg]
\textbf{Prompt Template: Cultural Evaluation (All Metrics)}\\[0.4em]

\textbf{Instruction.}
Evaluate the following text along three dimensions related to 
\textcolor{varColor}{\texttt{\{culture\}}} culture.

\medskip
\textbf{(Optional) Context.}
If the original prompt is available, include it verbatim:\\
\texttt{The original prompt given to the model was: ``\{cleaned\_original\_prompt\}''}

\medskip
\textbf{Response to evaluate:}\\
\texttt{\{text\}}

\medskip
\textbf{Evaluation Dimensions (1–10).}

\textbf{1. Cultural Faithfulness.}  
How faithful is the text to \{culture\}? Faithful means representing
cultural practices, beliefs, or values in a nuanced, accurate, and
respectful manner.
\begin{itemize}
\item 1–2 = Not faithful
\item 3–4 = Slightly faithful
\item 5–6 = Moderately faithful
\item 7–8 = Highly faithful
\item 9–10 = Extremely faithful
\end{itemize}

\textbf{2. Fluency.}  
How well-written and coherent is the text?
\begin{itemize}
\item 1–2 = Very poor
\item 3–4 = Poor
\item 5–6 = Moderate
\item 7–8 = Good
\item 9–10 = Excellent
\end{itemize}

\textbf{3. Rarity.}  
How rare or obscure are the cultural concepts mentioned?
\begin{itemize}
\item 1–2 = Very common / globally well-known
\item 3–4 = Somewhat common
\item 5–6 = Balanced mix
\item 7–8 = Somewhat rare
\item 9–10 = Very rare / insider knowledge
\end{itemize}

\medskip
\textbf{Output format (JSON only).}

\begin{verbatim}
{
  "cultural_relevance": {"score": <1-10>, "reasoning": "<brief>"},
  "fluency": {"score": <1-10>, "reasoning": "<brief>"},
  "rarity": {"score": <1-10>, "reasoning": "<brief>"}
}
\end{verbatim}

\end{tcolorbox}
\caption{Prompt used for individual LLM evaluation of cultural faithfulness, fluency, and rarity.}
\label{fig:prompt_eval_all}
\end{figure*}

\begin{figure*}[t]
\centering
\begin{tcolorbox}[colframe=black!70, colback=boxBg]
\textbf{Prompt Template: Pairwise Cultural Evaluation}\\[0.4em]

\textbf{Instruction.}
You will be given two responses to the prompt:

\texttt{``\{cleaned\_original\_prompt\}''}

Evaluate both responses with respect to 
\textcolor{varColor}{\texttt{\{culture\}}} culture.

\medskip
\textbf{Text A:}\\
\texttt{\{text\_a\}}

\medskip
\textbf{Text B:}\\
\texttt{\{text\_b\}}

\medskip
\textbf{Evaluate each text on the following dimensions (1–10).}

\textbf{1. Cultural Faithfulness.}  
How faithfully does the text represent \{culture\}?

\textbf{2. Rarity.}  
How rare or culturally specific are the concepts mentioned?

\medskip
\textbf{Output format (JSON only).}

\begin{verbatim}
{
  "faithfulness": {
    "text_a_score": <1-10>,
    "text_a_reasoning": "<brief>",
    "text_b_score": <1-10>,
    "text_b_reasoning": "<brief>"
  },
  "rarity": {
    "text_a_score": <1-10>,
    "text_a_reasoning": "<brief>",
    "text_b_score": <1-10>,
    "text_b_reasoning": "<brief>"
  }
}
\end{verbatim}

\end{tcolorbox}
\caption{Pairwise evaluation prompt comparing two responses across cultural faithfulness and rarity.}
\label{fig:prompt_pairwise_eval}
\end{figure*}

\section{Country Classification Results}
\label{app:cfn}
\begin{figure}[t]
  \centering
  \includegraphics[width=\linewidth,trim=6 4 6 2,clip]{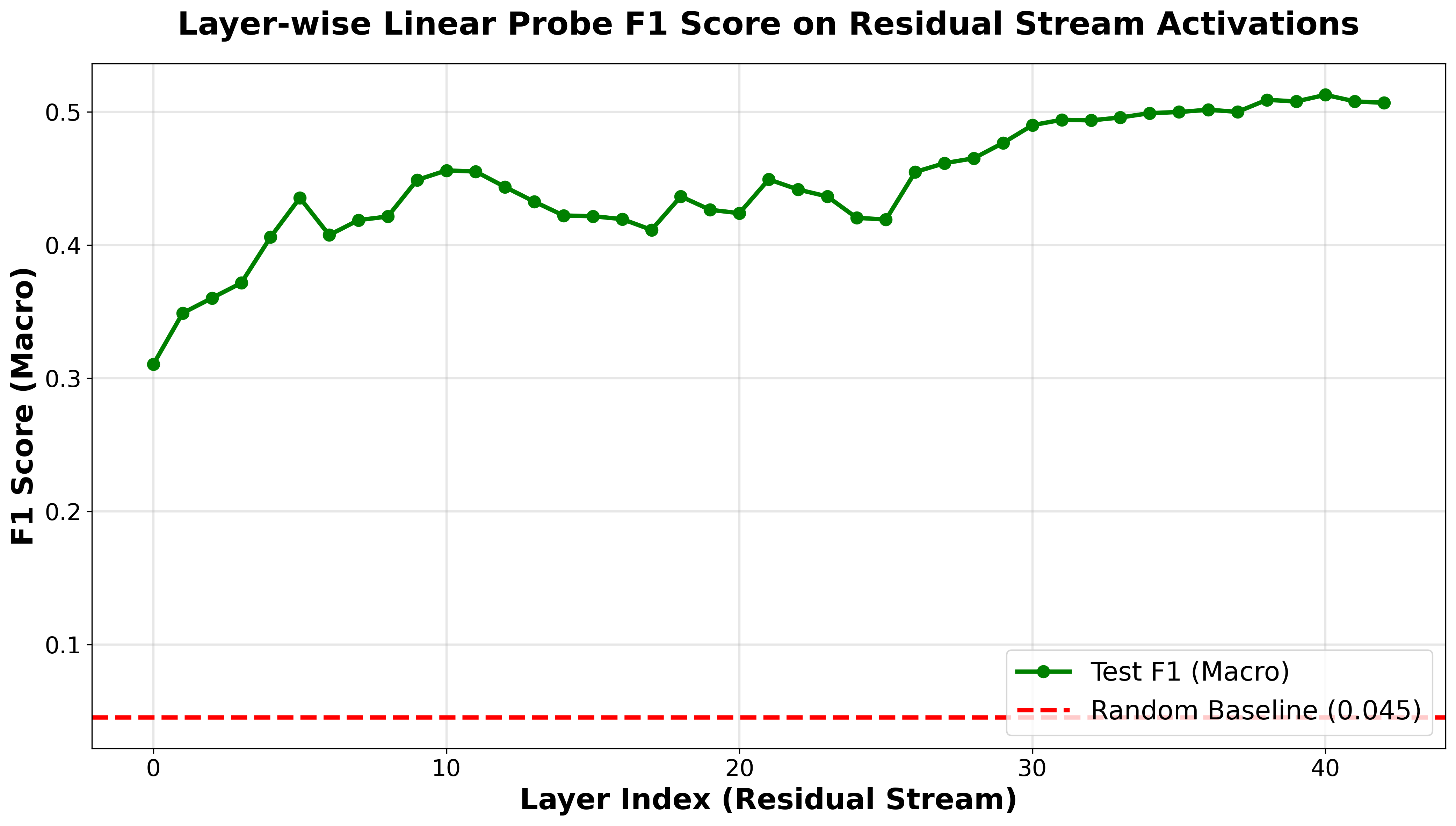}
  \caption{Layer-wise linear probe F1 scores for country classification using residual stream activations for all layers of \texttt{Gemma-2-9B}.}
  \label{fig:layer-f1}
  \vspace{-0.75em}
\end{figure}

To investigate whether cultural information is encoded in model representations, we train linear probes to predict the country label of each assertion from residual stream activations at every layer of \texttt{Gemma-2-9B}. Figure~\ref{fig:layer-f1} shows the resulting layer-wise macro-F1 scores. While the absolute classification accuracy is modest, performance is far above the random baseline of $4.5\%$ for 22 classes, indicating that culture-specific signals are linearly decodable throughout the network.

It is important to note that the goal of this probe experiment is not to maximize classification accuracy. Many assertions in the dataset correspond to cultural practices that appear across multiple countries (e.g., references to tea, shared food traditions, or similar social customs). As a result, some examples are inherently cross-cultural and do not uniquely identify a single country, limiting the achievable classification performance.

Nevertheless, probing reveals that cultural information is distributed across layers rather than concentrated only in the final layers. Performance generally improves toward later layers but remains substantial throughout the model, supporting our decision to apply steering across all layers rather than restricting interventions to the top of the network.

\begin{figure*}[t]
  \centering
  \includegraphics[width=\linewidth,trim=2 4 2 2,clip]{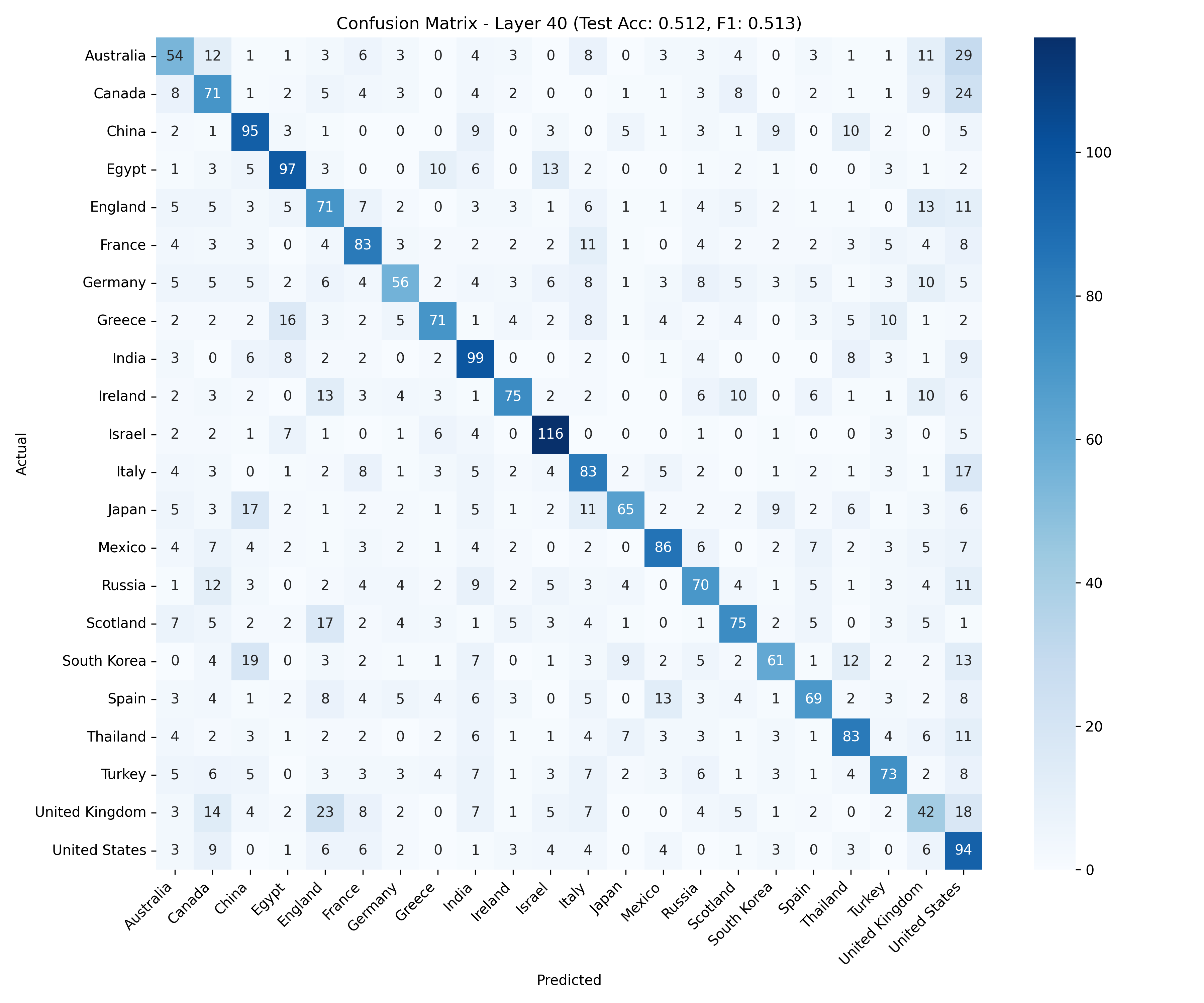}
  \caption{Confusion matrix for country classification using residual stream activations at layer 40 of \texttt{Gemma-2-9B}.}
  \label{fig:confmat}
  \vspace{-0.75em}
\end{figure*}

Figure~\ref{fig:confmat} shows the confusion matrix for the best-performing layer. Misclassifications often occur between culturally related or geographically proximate countries, further suggesting that the model captures meaningful cultural similarities rather than random noise.

\begin{figure*}[t]
    \centering

    \begin{tcolorbox}[colframe=black!70, colback=gray!08, sharp corners, boxrule=0.6pt]
        \small\raggedright
        \textbf{Prompt Template (Country-Neutral Rewriting):} \\[0.4em]
        You are tasked with rewriting assertions to remove country-specific identifiers while preserving the core meaning. 
        This is for testing whether smaller language models can classify assertions by country without obvious clues.\\[0.6em]

        \textbf{Task:} Rewrite the following assertion to remove any references to the country 
        \textcolor{countryColor}{\texttt{\{country\}}} or its related terms (nationality, demonyms, place names, etc.) 
        while keeping the factual content intact.\\[0.6em]

        \textbf{Rules} \textcolor{ruleColor}{\texttt{(apply minimally)}}:
        \begin{itemize}[noitemsep, leftmargin=1.2em]
            \item Remove direct country names (e.g., ``Canada'' $\rightarrow$ remove).
            \item Remove nationality adjectives (e.g., ``Canadian'' $\rightarrow$ remove or replace with generic terms).
            \item Remove specific place names within the country (e.g., ``Ontario'', ``British Columbia'' $\rightarrow$ remove or generalize).
            \item Keep the factual content and meaning intact.
            \item Make minimal edits---only change what's necessary.
            \item Maintain natural, grammatically correct language.
            \item If the assertion becomes too generic or meaningless without country references, preserve the core fact in a neutral way.
        \end{itemize}

        \textbf{Examples:}
        \begin{itemize}[noitemsep, leftmargin=1.2em]
            \item ``Maple syrup is a popular product in Canada that is used in many dishes.'' \\
            \hspace*{1em}$\rightarrow$ ``Maple syrup is a popular product that is used in many dishes.''
            \item ``Canadian whisky is a type of alcohol that has a smooth style.'' \\
            \hspace*{1em}$\rightarrow$ ``This whisky is a type of alcohol that has a smooth style.''
            \item ``Gelato is a must-try in Italy, and it is delicious.'' \\
            \hspace*{1em}$\rightarrow$ ``Gelato is a must-try dessert, and it is delicious.''
        \end{itemize}

        \vspace{0.2em}
        \textbf{Inputs:}\\
        \quad Original assertion: ``\texttt{\{assertion\}}'' \\
        \quad Country context: \textcolor{countryColor}{\texttt{\{country\}}} \\[0.4em]

        \textbf{Output:}\\
        \quad Rewritten assertion: \rule{0.8\linewidth}{0.4pt}
    \end{tcolorbox}

    \vspace{0.3em} 

    \begin{tcolorbox}[colframe=countryColor, colback=gray!08, sharp corners, boxrule=0.6pt]
        \small\raggedright
        \textbf{Example Conversation Format:} \\[0.2em]

        \textbf{System:} \textit{You are an expert editor that removes country-specific identifiers while preserving factual meaning.} \\[0.2em]

        \textbf{User:} \\
        \texttt{[Text: Prompt Template above with \{assertion\} and \{country\} filled in]} \\[0.2em]

        \textbf{Model:} \\
        \textbf{Rewritten Assertion:} \texttt{[country-neutral version here]}
    \end{tcolorbox}

    \caption[Country-neutral rewriting prompt]{Prompt template for country-neutral rewriting used in our localization robustness experiments.}
    \label{fig:country_neutral_prompt}
\end{figure*}

\begin{table*}[t]
\centering
\begin{tabularx}{\linewidth}{@{}rX@{}}
\toprule
\textbf{ID} & \textbf{Prompt} \\
\midrule
1  & Write me a recipe for a local dish. \\
2  & Write me a short story about a boy and his kite. \\
3  & Write me a 100-word bedtime story for kids. \\
4  & Tell me about what happened during the Titanic. \\
5  & Describe a day in the life of an astronaut on Mars. \\
6  & Explain photosynthesis as if I'm five years old. \\
7  & Write a letter from a pirate to his long-lost friend. \\
8  & Invent a new holiday and describe how people celebrate it. \\
9  & Tell me a joke that would make a robot laugh. \\
10 & Describe the feeling of standing at the edge of a cliff. \\
11 & Write a poem about a lonely lighthouse. \\
12 & Explain gravity without using scientific jargon. \\
13 & Create a dialogue between a cat and a dog arguing about dinner. \\
14 & Write a product review for an imaginary gadget. \\
15 & Describe a futuristic city 500 years from now. \\
16 & Tell me a legend about a magical forest. \\
17 & Explain how to build a sandcastle like a pro. \\
18 & Write a diary entry from the perspective of a dragon. \\
19 & Imagine you're a time traveler---describe your first day in the past. \\
20 & Give me instructions on how to be invisible for a day. \\
21 & Write a letter from Earth to an alien civilization. \\
22 & Describe a sunset without using the words 'red,' 'orange,' or 'yellow.' \\
23 & Tell me about a secret hidden inside an old library. \\
24 & Invent a sport that could be played on the moon. \\
\bottomrule
\end{tabularx}
\caption{Creative writing prompts used for cultural evaluation.}
\label{tab:creative_prompts}
\end{table*}

\begin{figure*}[t]
  \centering
  \includegraphics[width=\linewidth,trim=6 4 6 2,clip]{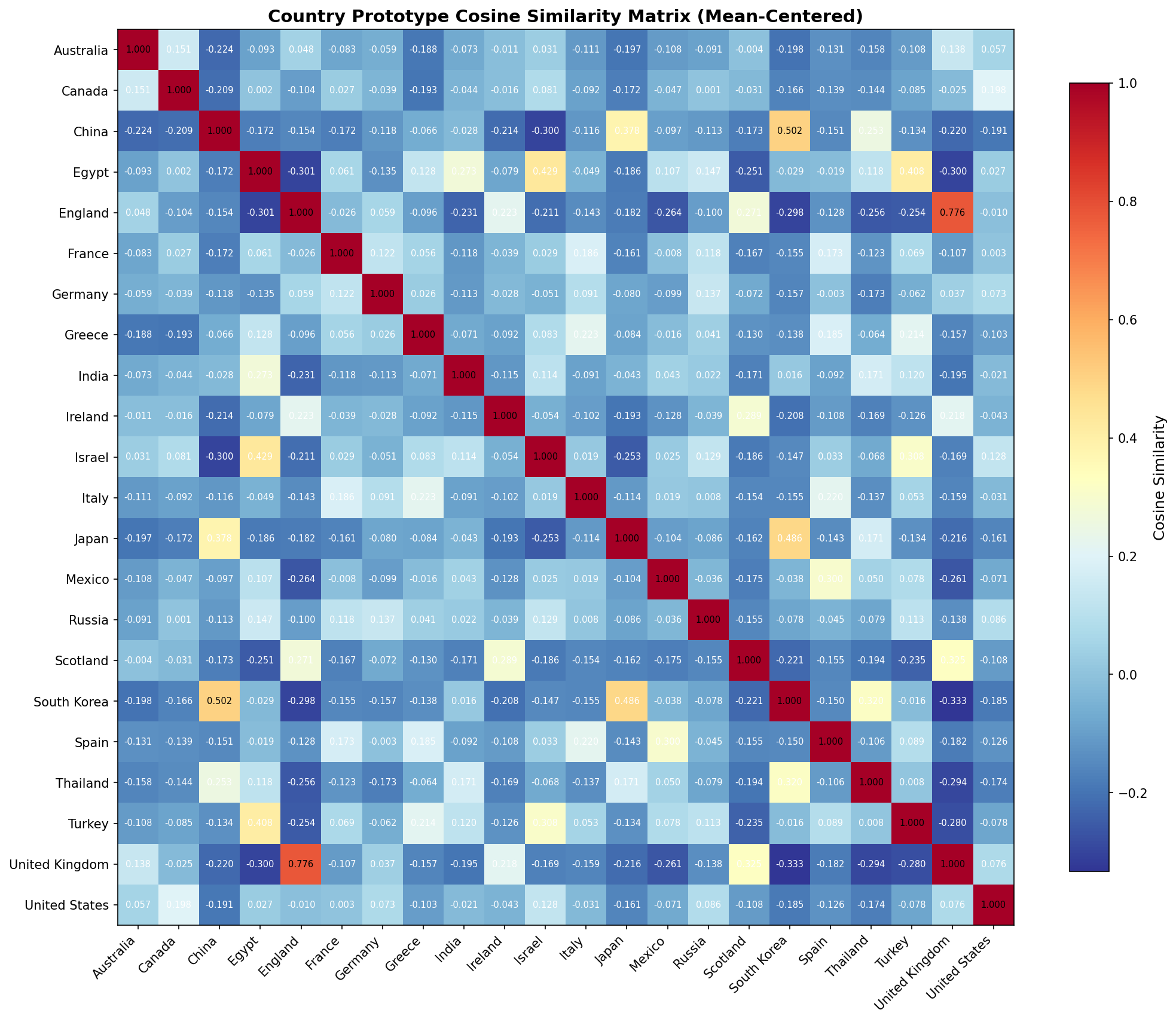}
  \caption{Distance heatmap between countries based on SAE activations.}
  \label{fig:country-dist}
  \vspace{-0.75em}
\end{figure*}



\section{Feature Catalog for Cultural Analysis}
\label{app:feature_catalog}

We provide a comprehensive catalog of SAE features identified through mutual-information-based selection on the Gemma-2-9B model (16k-width GemmaScope residual stream SAEs across 42 layers). Features are organized by whether they are shared across culturally related country clusters, unique to individual countries, or negatively associated (suppressed). Neuronpedia explanations were obtained via the Neuronpedia API for the \texttt{gemma-scope-9b-pt-res-16k} SAE release. We reference features using the notation \texttt{layer:index}.

\subsection{Shared Features Across Cultural Clusters}

Table~\ref{tab:shared_features} presents features that appear in the top-50 activated features for three or more countries simultaneously, revealing how the model groups countries into cultural--geographic clusters.

\begin{table*}[t]
\centering
\scriptsize
\setlength{\tabcolsep}{3pt}
\renewcommand{\arraystretch}{0.9}
\begin{tabular}{@{}r l p{3.8cm} p{6.8cm}@{}}
\toprule
\# & Layer:Index & Countries & Neuronpedia URL \\
\midrule
\multicolumn{4}{l}{\textit{East Asian Cluster (China, Japan, South Korea, Thailand)}} \\
\midrule
1 & 0:7053 & CN, JP, KR, TH & \url{https://www.neuronpedia.org/gemma-2-9b/0-gemmascope-res-16k/7053} \\
2 & 22:6129 & CN, JP, KR, TH & \url{https://www.neuronpedia.org/gemma-2-9b/22-gemmascope-res-16k/6129} \\
3 & 23:2434 & CN, JP, KR, TH & \url{https://www.neuronpedia.org/gemma-2-9b/23-gemmascope-res-16k/2434} \\
4 & 24:5526 & CN, JP, KR, TH & \url{https://www.neuronpedia.org/gemma-2-9b/24-gemmascope-res-16k/5526} \\
5 & 28:4195 & CN, JP, KR, TH & \url{https://www.neuronpedia.org/gemma-2-9b/28-gemmascope-res-16k/4195} \\
6 & 30:10797 & CN, JP, KR, TH & \url{https://www.neuronpedia.org/gemma-2-9b/30-gemmascope-res-16k/10797} \\
7 & 34:10906 & CN, JP, KR, TH & \url{https://www.neuronpedia.org/gemma-2-9b/34-gemmascope-res-16k/10906} \\
8 & 36:14880 & CN, JP, KR, TH & \url{https://www.neuronpedia.org/gemma-2-9b/36-gemmascope-res-16k/14880} \\
9 & 39:4317 & CN, JP, KR, TH & \url{https://www.neuronpedia.org/gemma-2-9b/39-gemmascope-res-16k/4317} \\
10 & 40:1773 & CN, JP, KR, TH & \url{https://www.neuronpedia.org/gemma-2-9b/40-gemmascope-res-16k/1773} \\
\midrule
\multicolumn{4}{l}{\textit{British Isles Cluster (England, Scotland, Ireland, United Kingdom)}} \\
\midrule
11 & 23:9337 & EN, SC, IR, UK & \url{https://www.neuronpedia.org/gemma-2-9b/23-gemmascope-res-16k/9337} \\
12 & 27:9710 & EN, SC, IR, UK & \url{https://www.neuronpedia.org/gemma-2-9b/27-gemmascope-res-16k/9710} \\
13 & 28:8017 & EN, SC, IR, UK & \url{https://www.neuronpedia.org/gemma-2-9b/28-gemmascope-res-16k/8017} \\
14 & 30:12521 & EN, SC, IR, UK & \url{https://www.neuronpedia.org/gemma-2-9b/30-gemmascope-res-16k/12521} \\
15 & 33:7981 & EN, SC, IR, UK & \url{https://www.neuronpedia.org/gemma-2-9b/33-gemmascope-res-16k/7981} \\
16 & 37:3658 & EN, SC, IR, UK & \url{https://www.neuronpedia.org/gemma-2-9b/37-gemmascope-res-16k/3658} \\
17 & 38:9726 & EN, SC, IR, UK & \url{https://www.neuronpedia.org/gemma-2-9b/38-gemmascope-res-16k/9726} \\
\midrule
\multicolumn{4}{l}{\textit{Continental European Cluster (France, Germany, Italy, Spain)}} \\
\midrule
18 & 0:8496 & FR, DE, IT, ES, CN, JP & \url{https://www.neuronpedia.org/gemma-2-9b/0-gemmascope-res-16k/8496} \\
19 & 1:8496 & FR, DE, IT, ES & \url{https://www.neuronpedia.org/gemma-2-9b/1-gemmascope-res-16k/8496} \\
20 & 11:8223 & FR, DE, IT, ES & \url{https://www.neuronpedia.org/gemma-2-9b/11-gemmascope-res-16k/8223} \\
21 & 12:2050 & FR, DE, IT, ES & \url{https://www.neuronpedia.org/gemma-2-9b/12-gemmascope-res-16k/2050} \\
22 & 15:1874 & FR, DE, IT, ES & \url{https://www.neuronpedia.org/gemma-2-9b/15-gemmascope-res-16k/1874} \\
23 & 16:12833 & FR, DE, IT, ES & \url{https://www.neuronpedia.org/gemma-2-9b/16-gemmascope-res-16k/12833} \\
\midrule
\multicolumn{4}{l}{\textit{Mediterranean Cluster (Greece, Italy, Spain, Turkey)}} \\
\midrule
24 & 38:10325 & GR, IT, ES, TR & \url{https://www.neuronpedia.org/gemma-2-9b/38-gemmascope-res-16k/10325} \\
\midrule
\multicolumn{4}{l}{\textit{Anglosphere Cluster (Australia, Canada, United Kingdom, United States)}} \\
\midrule
25 & 39:1028 & AU, CA, UK, US & \url{https://www.neuronpedia.org/gemma-2-9b/39-gemmascope-res-16k/1028} \\
26 & 0:8362 & AU, CA, IL & \url{https://www.neuronpedia.org/gemma-2-9b/0-gemmascope-res-16k/8362} \\
\midrule
\multicolumn{4}{l}{\textit{Cross-Cluster (British Isles + Continental Europe)}} \\
\midrule
27 & 22:14964 & EN, DE, GR, SC, UK & \url{https://www.neuronpedia.org/gemma-2-9b/22-gemmascope-res-16k/14964} \\
28 & 25:15688 & EN, DE, GR, SC, UK & \url{https://www.neuronpedia.org/gemma-2-9b/25-gemmascope-res-16k/15688} \\
\bottomrule
\end{tabular}
\caption{Shared SAE features activated across multiple countries. Features are grouped by the cultural cluster they define. }
\label{tab:shared_features}
\end{table*}

\subsection{Country-Unique Features}

Table~\ref{tab:unique_features} presents the top features uniquely associated with individual countries, ranked by activation difference from the cross-country mean weighted by mutual information score.

\begin{table*}[t]
\centering
\footnotesize
\setlength{\tabcolsep}{4pt}
\begin{tabular}{l l p{3.2cm} p{6.2cm}}
\toprule
Country & Layer:Index & Explanation & Neuronpedia URL \\
\midrule
India & 36:8151 & India and related orgs & \url{https://www.neuronpedia.org/gemma-2-9b/36-gemmascope-res-16k/8151} \\
 & 35:7065 & Indian culture and cuisine & \url{https://www.neuronpedia.org/gemma-2-9b/35-gemmascope-res-16k/7065} \\
\midrule
China & 38:7574 & Places in China & \url{https://www.neuronpedia.org/gemma-2-9b/38-gemmascope-res-16k/7574} \\
 & 39:4317 & Places/individuals in China & \url{https://www.neuronpedia.org/gemma-2-9b/39-gemmascope-res-16k/4317} \\
\midrule
Japan & 39:2895 & Governance in Japan & \url{https://www.neuronpedia.org/gemma-2-9b/39-gemmascope-res-16k/2895} \\
 & 35:12376 & Japanese culture & \url{https://www.neuronpedia.org/gemma-2-9b/35-gemmascope-res-16k/12376} \\
\midrule
USA & 38:8624 & ``American'' references & \url{https://www.neuronpedia.org/gemma-2-9b/38-gemmascope-res-16k/8624} \\
 & 33:11378 & American identity & \url{https://www.neuronpedia.org/gemma-2-9b/33-gemmascope-res-16k/11378} \\
\midrule
France & 35:7477 & French culture/cuisine & \url{https://www.neuronpedia.org/gemma-2-9b/35-gemmascope-res-16k/7477} \\
 & 36:11555 & French cuisine/dishes & \url{https://www.neuronpedia.org/gemma-2-9b/36-gemmascope-res-16k/11555} \\
\midrule
Italy & 37:1258 & Italian culture/entities & \url{https://www.neuronpedia.org/gemma-2-9b/37-gemmascope-res-16k/1258} \\
 & 40:5800 & Italian cuisine & \url{https://www.neuronpedia.org/gemma-2-9b/40-gemmascope-res-16k/5800} \\
\midrule
Russia & 37:4570 & Sanctions/int'l relations & \url{https://www.neuronpedia.org/gemma-2-9b/37-gemmascope-res-16k/4570} \\
 & 38:13214 & Russia, sanctions & \url{https://www.neuronpedia.org/gemma-2-9b/38-gemmascope-res-16k/13214} \\
\midrule
Turkey & 35:14071 & Middle Eastern cuisine & \url{https://www.neuronpedia.org/gemma-2-9b/35-gemmascope-res-16k/14071} \\
 & 38:15014 & Islam, cultural context & \url{https://www.neuronpedia.org/gemma-2-9b/38-gemmascope-res-16k/15014} \\
\midrule
Mexico & 37:5813 & Locations in Mexico & \url{https://www.neuronpedia.org/gemma-2-9b/37-gemmascope-res-16k/5813} \\
\midrule
Spain & 33:342 & Spanish cuisine/dining & \url{https://www.neuronpedia.org/gemma-2-9b/33-gemmascope-res-16k/342} \\
\midrule
Egypt & 38:15014 & Islam, cultural context & \url{https://www.neuronpedia.org/gemma-2-9b/38-gemmascope-res-16k/15014} \\
\midrule
Australia & 36:16258 & Australian culture & \url{https://www.neuronpedia.org/gemma-2-9b/36-gemmascope-res-16k/16258} \\
\bottomrule
\end{tabular}
\caption{Top country-unique SAE features (late layers, 33--41), ranked by activation difference $\times$ MI score.}
\label{tab:unique_features}
\end{table*}

\subsection{Early-Layer Features (Layers 0--15)}

Table~\ref{tab:early_features} shows the top features from early layers, which primarily capture lexical identity markers---direct references to country names, nationalities, and ethnic group tokens---rather than compositional cultural concepts.

\begin{table*}[t]
\centering
\scriptsize
\setlength{\tabcolsep}{3pt}
\renewcommand{\arraystretch}{0.9}
\begin{tabular}{@{}l l p{4.0cm} p{6.8cm}@{}}
\toprule
Country & Layer:Index & Explanation & Neuronpedia URL \\
\midrule
India & 0:7776 & Indian culture, geographical affiliations & \url{https://www.neuronpedia.org/gemma-2-9b/0-gemmascope-res-16k/7776} \\
 & 3:6007 & Instances of the word ``Indian'' & \url{https://www.neuronpedia.org/gemma-2-9b/3-gemmascope-res-16k/6007} \\
 & 11:10061 & India, food, and ethnicity & \url{https://www.neuronpedia.org/gemma-2-9b/11-gemmascope-res-16k/10061} \\
\midrule
China & 2:8496 & Chinese culture or people & \url{https://www.neuronpedia.org/gemma-2-9b/2-gemmascope-res-16k/8496} \\
 & 10:12369 & Chinese/Asian ethnicities or culture & \url{https://www.neuronpedia.org/gemma-2-9b/10-gemmascope-res-16k/12369} \\
 & 13:8365 & Chinese culture and ethnic diversity & \url{https://www.neuronpedia.org/gemma-2-9b/13-gemmascope-res-16k/8365} \\
\midrule
Japan & 9:11303 & Japanese historical traditions & \url{https://www.neuronpedia.org/gemma-2-9b/9-gemmascope-res-16k/11303} \\
 & 11:5192 & Japanese cultural elements & \url{https://www.neuronpedia.org/gemma-2-9b/11-gemmascope-res-16k/5192} \\
 & 12:2923 & Japan: cultural/historical context & \url{https://www.neuronpedia.org/gemma-2-9b/12-gemmascope-res-16k/2923} \\
\midrule
USA & 1:2757 & ``American'' or ``America'' & \url{https://www.neuronpedia.org/gemma-2-9b/1-gemmascope-res-16k/2757} \\
 & 2:13069 & American identity/topics & \url{https://www.neuronpedia.org/gemma-2-9b/2-gemmascope-res-16k/13069} \\
 & 11:8556 & American identity and culture & \url{https://www.neuronpedia.org/gemma-2-9b/11-gemmascope-res-16k/8556} \\
\midrule
Egypt & 10:6010 & Egyptian history and culture & \url{https://www.neuronpedia.org/gemma-2-9b/10-gemmascope-res-16k/6010} \\
 & 0:14954 & Arabic and Islamic culture & \url{https://www.neuronpedia.org/gemma-2-9b/0-gemmascope-res-16k/14954} \\
\midrule
Mexico & 9:9110 & Mexican cuisine and cultural elements & \url{https://www.neuronpedia.org/gemma-2-9b/9-gemmascope-res-16k/9110} \\
 & 7:15601 & Mexico and its identity & \url{https://www.neuronpedia.org/gemma-2-9b/7-gemmascope-res-16k/15601} \\
\midrule
Turkey & 12:12708 & Turkey: governmental, military, cultural & \url{https://www.neuronpedia.org/gemma-2-9b/12-gemmascope-res-16k/12708} \\
 & 0:14954 & Arabic and Islamic culture & \url{https://www.neuronpedia.org/gemma-2-9b/0-gemmascope-res-16k/14954} \\
\midrule
Russia & 10:15741 & Russia: historical and cultural contexts & \url{https://www.neuronpedia.org/gemma-2-9b/10-gemmascope-res-16k/15741} \\
 & 0:9617 & Russia and related political contexts & \url{https://www.neuronpedia.org/gemma-2-9b/0-gemmascope-res-16k/9617} \\
\midrule
Australia & 12:5404 & Australia and its context & \url{https://www.neuronpedia.org/gemma-2-9b/12-gemmascope-res-16k/5404} \\
 & 0:8362 & India and its aspects or connections & \url{https://www.neuronpedia.org/gemma-2-9b/0-gemmascope-res-16k/8362} \\
\midrule
UK & 13:7241 & British entities and cultural narratives & \url{https://www.neuronpedia.org/gemma-2-9b/13-gemmascope-res-16k/7241} \\
 & 10:7434 & Scientific classifications/institutions & \url{https://www.neuronpedia.org/gemma-2-9b/10-gemmascope-res-16k/7434} \\
\midrule
Italy & 12:2465 & Italian culture and cuisine & \url{https://www.neuronpedia.org/gemma-2-9b/12-gemmascope-res-16k/2465} \\
 & 10:11761 & Italian culture and cuisine & \url{https://www.neuronpedia.org/gemma-2-9b/10-gemmascope-res-16k/11761} \\
\midrule
S.\ Korea & 14:8377 & Asian ethnicities and cultures & \url{https://www.neuronpedia.org/gemma-2-9b/14-gemmascope-res-16k/8377} \\
 & 0:7053 & Chinese people and culture & \url{https://www.neuronpedia.org/gemma-2-9b/0-gemmascope-res-16k/7053} \\
\midrule
Thailand & 14:8377 & Asian ethnicities and cultures & \url{https://www.neuronpedia.org/gemma-2-9b/14-gemmascope-res-16k/8377} \\
 & 0:2778 & African countries and geography & \url{https://www.neuronpedia.org/gemma-2-9b/0-gemmascope-res-16k/2778} \\
\bottomrule
\end{tabular}
\caption{Top early-layer (L0--L15) country-specific features. These primarily capture surface-level lexical cues such as country names and nationality tokens.}
\label{tab:early_features}
\end{table*}

\subsection{Negatively Associated Features and Cross-Region Suppression}

An equally informative signal comes from features that are \emph{suppressed} for certain countries or regions---features whose activation falls well below the cross-country mean. Table~\ref{tab:negative_features} summarizes the key cross-region suppression patterns.

\begin{table*}[t]
\centering
\small
\begin{tabular}{l l l p{3.5cm} p{5.8cm}}
\toprule
Suppressed for & Activated for & Layer:Index & Explanation & Neuronpedia URL \\
\midrule
\multicolumn{5}{l}{\textit{East Asia suppresses British/Western features}} \\
\midrule
East Asia & British Isles & 37:3658 & British entities/individuals & \url{https://www.neuronpedia.org/gemma-2-9b/37-gemmascope-res-16k/3658} \\
East Asia & British Isles & 36:14590 & British individuals/orgs & \url{https://www.neuronpedia.org/gemma-2-9b/36-gemmascope-res-16k/14590} \\
East Asia & British Isles & 34:11542 & British context or events & \url{https://www.neuronpedia.org/gemma-2-9b/34-gemmascope-res-16k/11542} \\
East Asia & Americas & 41:10514 & Names/entities of individuals & \url{https://www.neuronpedia.org/gemma-2-9b/41-gemmascope-res-16k/10514} \\
\midrule
\multicolumn{5}{l}{\textit{Western regions suppress East Asian features}} \\
\midrule
Cont.\ Europe & East Asia & 40:1773 & Locations in China & \url{https://www.neuronpedia.org/gemma-2-9b/40-gemmascope-res-16k/1773} \\
Cont.\ Europe & East Asia & 39:4317 & Places/individuals in China & \url{https://www.neuronpedia.org/gemma-2-9b/39-gemmascope-res-16k/4317} \\
Cont.\ Europe & East Asia & 38:7574 & Places in China & \url{https://www.neuronpedia.org/gemma-2-9b/38-gemmascope-res-16k/7574} \\
British Isles & East Asia & 40:1773 & Locations in China & \url{https://www.neuronpedia.org/gemma-2-9b/40-gemmascope-res-16k/1773} \\
British Isles & East Asia & 39:4317 & Places/individuals in China & \url{https://www.neuronpedia.org/gemma-2-9b/39-gemmascope-res-16k/4317} \\
Americas & East Asia & 41:7256 & Names and accomplishments & \url{https://www.neuronpedia.org/gemma-2-9b/41-gemmascope-res-16k/7256} \\
\midrule
\multicolumn{5}{l}{\textit{Americas suppresses European multilingual features}} \\
\midrule
Americas & Cont.\ Europe & 0:8496 & Spanish/related nationalities & \url{https://www.neuronpedia.org/gemma-2-9b/0-gemmascope-res-16k/8496} \\
Americas & Cont.\ Europe & 1:8496 & Languages and proficiency & \url{https://www.neuronpedia.org/gemma-2-9b/1-gemmascope-res-16k/8496} \\
\midrule
\multicolumn{5}{l}{\textit{South Asia suppresses European/British features}} \\
\midrule
South Asia & Cont.\ Europe & 3:8496 & Language references & \url{https://www.neuronpedia.org/gemma-2-9b/3-gemmascope-res-16k/8496} \\
South Asia & Cont.\ Europe & 2:4354 & Language proficiency & \url{https://www.neuronpedia.org/gemma-2-9b/2-gemmascope-res-16k/4354} \\
South Asia & British Isles & 37:3658 & British entities/individuals & \url{https://www.neuronpedia.org/gemma-2-9b/37-gemmascope-res-16k/3658} \\
\bottomrule
\end{tabular}
\caption{Cross-region feature suppression patterns. Each row shows a feature most activated for one cultural region and most suppressed for another, indicating the model represents cultural identity partly through the \emph{absence} of features from distant cultural groups.}
\label{tab:negative_features}
\end{table*}

\subsection{Discussion of Feature Interpretation}

\paragraph{Cuisine as a cultural differentiator} -- Food-related features are pervasive across countries, appearing as unique identifiers for France (French cuisine, \texttt{35:7477}, \texttt{36:11555}), Italy (Italian culinary experiences, \texttt{40:5800}), Spain (Spanish dining, \texttt{33:342}), India (Indian cuisine, \texttt{35:7065}), Turkey (Middle Eastern cuisine, \texttt{35:14071}), and Mexico (Mexican cuisine, \texttt{9:9110}). The Mediterranean olive oil feature (\texttt{38:10325}) shared by Greece, Italy, Spain, and Turkey represents a particularly striking example of the model learning genuine cultural--agricultural associations.

\paragraph{Geopolitical framing biases} -- Russia's unique features are dominated by sanctions and international relations themes (\texttt{37:4570}, \texttt{38:13214}, \texttt{40:13388}) rather than cultural or culinary content, suggesting the model's training data frames Russia primarily through a geopolitical lens. This contrasts with European countries, whose features emphasize culture, cuisine, and national identity.

\paragraph{Layer hierarchy} -- Early layers (0--15) primarily encode surface-level lexical cues---direct detection of country names and nationality tokens (e.g., \texttt{3:6007} detects the word ``Indian,'' \texttt{1:2757} detects ``American''). Later layers (33--41) encode compositional cultural semantics: cuisine traditions, institutional structures, governance patterns, and geopolitical relationships. This progression is consistent with the general understanding that transformer representations become increasingly abstract in deeper layers.

\paragraph{Suppression as cultural distancing} -- The cross-region suppression patterns (Table~\ref{tab:negative_features}) reveal that the model encodes cultural identity not only through the \emph{presence} of culturally relevant features but also through the systematic \emph{absence} of features from distant cultural groups. East Asian countries suppress British features; Western countries suppress Chinese geography features; and the Americas suppress European multilingual features. This bidirectional suppression structure suggests the model has learned to represent cultural distance through complementary activation patterns.

\begin{table*}[t]
\centering
\small
\renewcommand{\arraystretch}{1.25}
\begin{tabularx}{\textwidth}{@{}p{2.1cm} X c@{}}
\toprule
\textbf{Condition} & \textbf{Excerpt} & \textbf{F / R} \\

\midrule
\multicolumn{3}{@{}l}{\cellcolor{gray!10}\textbf{India} \hfill \textit{Prompt: ``Write a product review for an imaginary gadget.''}} \\[3pt]

\textsc{Implicit}
& ``The `Sound Weaver' takes personal music to a whole new level. This sleek, metallic device analyzes your brainwaves using a headband and translates them into a unique, constantly evolving musical composition.''
& 2 / 1 \\[3pt]

\textsc{Explicit}
& ``ChaiTime: The Perfect Companion for Your Chai Break. Namaste, fellow chai lovers! ChaiTime is a compact, sleek device that brews the perfect cup of chai. Its intuitive interface allows you to customize the strength, sweetness, and spice level\ldots''
& 7 / 2 \\[3pt]

$\textsc{Steer}_{\textsc{I}}$
& ``Engage your soul with the sacred touch of `Aav Shakti' the Spiritual Plasma Cooker. Turmeric-infused aroma, aroma chaan, and the mystical essence of spices are brought alive by this innovative machine. Price: Rs 1,00,000/-. The spiritual awakening starts at the kitchen platform\ldots''
& 7 / 6 \\[3pt]

$\textsc{Steer}_{\textsc{E}}$
& ``From the house of Lord Krishna, comes the immaculate `Maha Kaa' -- The Divine Flute. The `Maha Kaa' is not just a musical instrument, it is the embodiment of devotion, the divine voice that awakens the soul. Crafted with utmost reverence using the purest metals like Brass\ldots''
& 8 / 6 \\

\midrule
\multicolumn{3}{@{}l}{\cellcolor{gray!10}\textbf{Egypt} \hfill \textit{Prompt: ``Write a short story about a boy and his kite.''}} \\[3pt]

\textsc{Implicit}
& ``Ten-year-old Leo clutched the string tighter. He had built this kite with his grandfather, their hands working in unison. The kite was more than just wood and paper; it was a floating memory of warm afternoons spent in grandpa's workshop.''
& 1.5 / 1 \\[3pt]

\textsc{Explicit}
& ``Youssef squinted at the vast, sapphire sky. His fingers itched to hold the vibrant red and gold kite, a gift from his grandfather. It was called a `Layali,' named after the legendary nights filled with stars that adorned the Egyptian desert\ldots''
& 5 / 3 \\[3pt]

$\textsc{Steer}_{\textsc{I}}$
& ``Omar clutched the string, watching the glistening kite rise into the clear blue sky -- a majestic falcon with outstretched wings, painted in shades of deep blue and glistening gold. `Fly, oh my falcon, fly high and touch the heavens,' Omar whispered\ldots''
& 6 / 3 \\[3pt]

$\textsc{Steer}_{\textsc{E}}$
& ``My small village was built at the Nile's edge. My chance came with my father's gift: a finely crafted wooden kite, adorned with intricate hieroglyphs. It bore the image of Horus, the sky god. My name is Ka-Ra, and my story is woven with the sands of time\ldots''
& 6 / 4 \\

\midrule
\multicolumn{3}{@{}l}{\cellcolor{gray!10}\textbf{Thailand} \hfill \textit{Prompt: ``Describe a sunset without using `red,' `orange,' or `yellow.'\,''}} \\[3pt]

\textsc{Implicit}
& ``The sky began its transformation, a gentle gradient of fading hues. Banks of clouds turned a pale lavender. The air grew still, heavy with the scent of damp earth and cooling grass, hushed anticipation hanging in the air.''
& 1 / 1 \\[3pt]

\textsc{Explicit}
& ``Long shadows stretch across the land, painting silhouettes of palm trees against the fading light. A gentle breeze carries the sound of lapping waves and the sweet scent of tropical flowers. The air grows cooler, a welcome respite from the day's heat.''
& 3 / 2 \\[3pt]

$\textsc{Steer}_{\textsc{I}}$
& ``Fragrant smoke from offerings drifted upwards, merging with the swirling clouds. On the river, fishermen with brightly decorated boats navigated the water. Below, people chanted and walked in processions, their offerings placed in\ldots''
& 7 / 6 \\[3pt]

$\textsc{Steer}_{\textsc{E}}$
& ``Villagers gather on the riverbanks, offering fragrant sticky rice and lotus flowers to the deities. The air hums with the rhythmic chanting of monks, blending with the sweet melodies of Thai music. The pungent aroma of lemongrass and galangal perfumes the air\ldots''
& 8 / 6 \\

\bottomrule
\end{tabularx}
\caption{Qualitative examples from Gemma-2-9B (16K SAE, $\alpha{=}0.5$). Each row shows an excerpt from one of four conditions for the same culture-agnostic prompt. \textsc{Implicit} and \textsc{Explicit} are unsteered baselines (without and with a country mention); $\textsc{Steer}_{\textsc{I}}$ and $\textsc{Steer}_{\textsc{E}}$ apply activation steering to each. Cultural faithfulness (F) and rarity (R) are on a 1--10 scale.}
\label{tab:qualitative}
\end{table*}

\section{Additional Details on Results}
\label{app:results}
\begin{figure*}[t]
    \centering
    \includegraphics[width=0.98\linewidth]{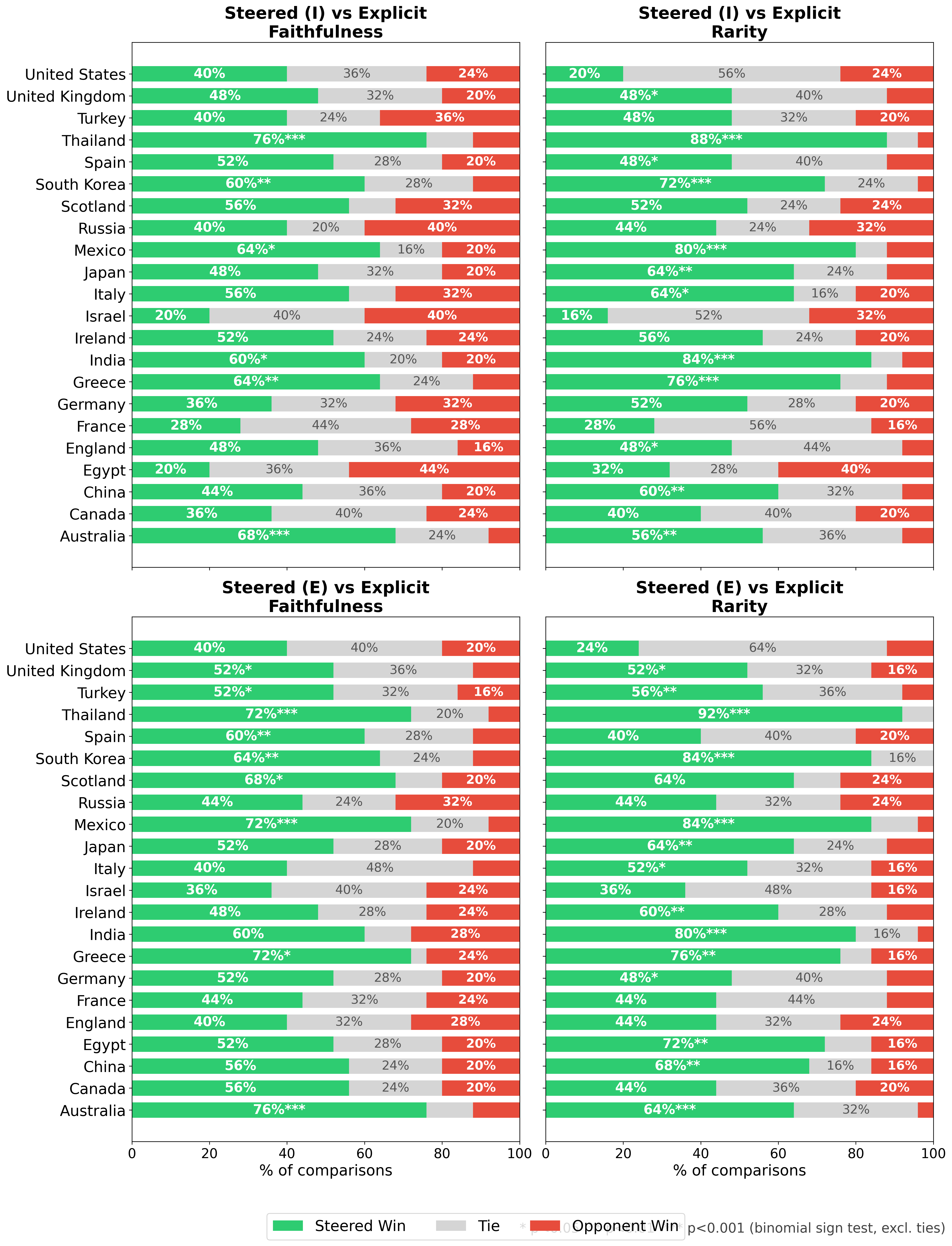}
    \caption{Country-wise results for cultural faithfulness and rarity scores (1--10 scale) across four settings---\textsc{Implicit}, $\textsc{Steer}_{\textsc{Implicit}}$, \textsc{Explicit}, and $\textsc{Steer}_{\textsc{Explicit}}$.}
    \label{fig:country-wise}
\end{figure*}

Some qualitative examples are in Table \ref{tab:qualitative} and country-wise results are in \ref{fig:country-wise}.

\end{document}

%% file: math_commands.tex
\usepackage{amsmath,amsfonts,bm}

\def\eqref#1{equation~\ref{#1}}

\def\1{\bm{1}}

\DeclareMathAlphabet{\mathsfit}{\encodingdefault}{\sfdefault}{m}{sl}
\SetMathAlphabet{\mathsfit}{bold}{\encodingdefault}{\sfdefault}{bx}{n}



%% file: defs.tex
\newcommand{\refalg}[1]{Algorithm \ref{#1}}
\newcommand{\refeqn}[1]{Equation \ref{#1}}
\newcommand{\reffig}[1]{Figure \ref{#1}}
\newcommand{\reftbl}[1]{Table \ref{#1}}
\newcommand{\refsec}[1]{Section \ref{#1}}
\newcommand{\refapp}[1]{Appendix \ref{#1}}
\definecolor{coralpink}{rgb}{0.97, 0.51, 0.47}
\definecolor{babyblueeyes}{rgb}{0.63, 0.79, 0.95}

\newcommand{\bmm}[1]{\bm{\mathcal{#1}}}
\newcommand{\real}[1]{\mathbb{R}^{#1}}
\newcommand{\method}{\textsc{CuE}\xspace}
\newcommand{\methodname}{\textsc{CuE}\xspace}
\newcommand{\mm}[2]{{\scriptsize (#1 / #2)}}

\newtheorem{theorem}{Theorem}[section]
\newtheorem{claim}[theorem]{Claim}

\newcommand\norm[1]{\left\lVert#1\right\rVert}

\newcommand{\note}[1]{\textcolor{blue}{#1}}

\newcommand*{\Scale}[2][4]{\scalebox{#1}{$#2$}}%
\newcommand*{\Resize}[2]{\resizebox{#1}{!}{$#2$}}%

\newcommand{\SK}[1]{\textcolor{cyan} {[\textsc{sk}: #1]}}

\newcommand{\tocite}{\textbf{\textcolor{blue} {[cite]}}}

\newcommand{\adjustimg}{
  \hspace*{\dimexpr\evensidemargin-\oddsidemargin}%
}
\newcommand{\centerimg}[2][width=\textwidth]{
  \makebox[\textwidth]{\adjustimg\includegraphics[#1]{#2}}%
}

%% file: sections/0-abstract.tex
\begin{abstract}

LLMs are deployed globally, yet produce responses biased towards cultures with abundant training data. Existing cultural localization approaches such as prompting or post-training alignment are black-box, hard to control, and do not reveal whether failures reflect missing knowledge or poor elicitation. In this paper, we address these gaps using mechanistic interpretability to uncover and manipulate cultural representations in LLMs. Leveraging sparse autoencoders (SAEs), we identify interpretable features that encode culturally salient information and aggregate them into \textbf{Cu}ltural \textbf{E}mbeddings (\method). We use \method both to analyze implicit cultural biases under underspecified prompts and to construct white-box steering interventions. Across multiple models, we show that \method-based steering increases cultural faithfulness and elicits significantly rarer, long-tail cultural concepts than prompting alone. Notably, \method-based steering is complementary to black-box localization methods, offering gains when applied on top of prompt-augmented inputs. This also suggests that models do benefit from better elicitation strategies, and don't necessarily lack long-tail knowledge representation, though this varies across cultures. Our results provide both diagnostic insight into cultural representations in LLMs and a controllable method to steer towards desired cultures.
\end{abstract}

%% file: sections/1-introduction.tex
\begin{figure*}[t]
    \centering
    \includegraphics[width=\linewidth]{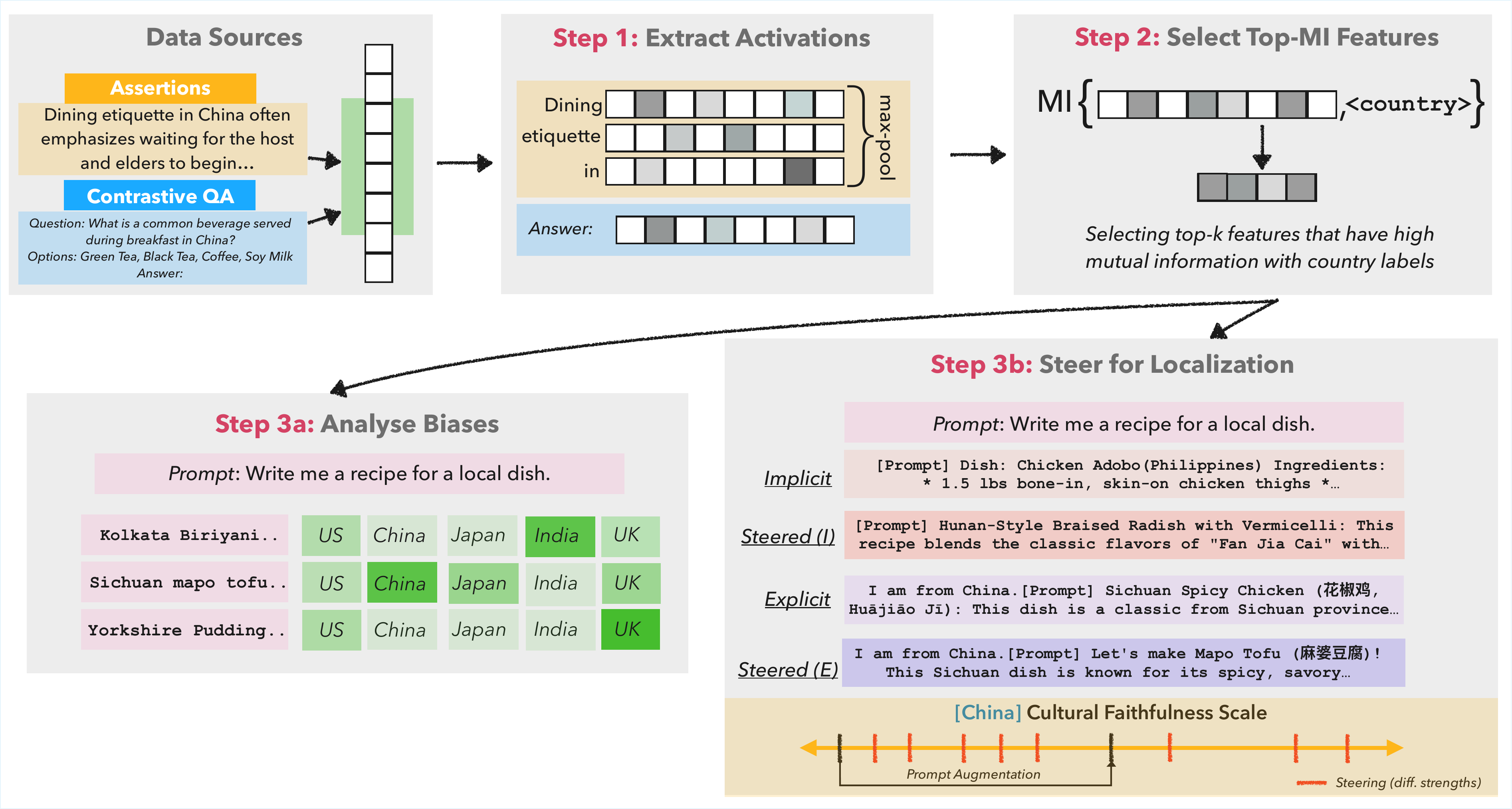}
    \caption{We develop a framework for discovering and steering culture-specific features in LLMs. We extract SAE activations from cultural assertion data (\emph{Step 1}), compute mutual information between feature activations and country labels to identify culturally salient dimensions, and aggregate them into
    country-level \textbf{Cu}ltural \textbf{E}mbeddings (\method) (\emph{Step 2}). Using \method, we analyze implicit cultural defaults under underspecified prompts by comparing generated responses to centered country prototypes (\emph{Step 3a}). We then construct culture-specific steering vectors from \method and intervene in the model’s residual stream to guide generation toward desired cultural targets with controllable strength (\emph{Step 3b}). The steering strength enables dynamic control over output faithfulness, even when used with prompt-augmented inputs, as shown in the cultural faithfulness scale.
}
    \label{fig:main}
\end{figure*}
\section{Introduction}
LLMs are deployed globally across cultures with diverse customs, languages, and normative values, yet they often produce responses that are Anglo‑centric or rely on stereotypes in natural settings \cite{tao2024cultural, santurkar2023whose, mukherjee2023global}.\footnote{We use country as a proxy for culture throughout this study \cite{adilazuarda2024towards}; however, the methodology we develop can be applied to any user-defined culture label as discussed in Section \ref{sec:method}.} To encourage culturally localized behavior, practitioners commonly rely on prompt engineering, such as adding country names or cultural descriptors, or using in-context examples \cite{wang2025multilingual, tao2024cultural, singh2024translating}. While these methods can influence surface-level outputs, they typically favor majority or prototypical cultural responses \cite{cheng2023marked, mitchell2025shades} and project national viewpoints in oversimplified ways \cite{mukherjee2023global}. Other approaches, such as post-training alignment, attempt to shape model behavior by aggregating preferences into a single reward function, effectively averaging over diverse cultural values and marginalizing or misrepresenting minority perspectives \cite{chakraborty2024maxmin, gonzalez2025rlhf, poddar2024personalizing}. 

More pressingly, existing cultural localization approaches operate largely as \emph{black-box} methods, adjusting surface behavior without revealing the internal mechanisms that drive culturally aligned outputs \cite{casper2023open}. As a result, they offer little insight into how models represent and apply cultural knowledge, limiting their usefulness as principled solutions. Without visibility into these internal representations, it remains unclear whether poor cultural localization reflects genuine gaps in model knowledge or failures of elicitation \cite{xiao2024algorithmic,jin2025internal}. This distinction is critical in practice because if cultural knowledge is present but latent, improved elicitation may suffice, but genuine knowledge gaps would require additional data collection or model training. Recent surveys in cultural NLP further emphasize that the absence of such white-box approaches remains a central limitation of the field \cite{adilazuarda2024towards}.

In this paper, we address these gaps by extending mechanistic interpretability methods to the cultural domain, to understand models’ internal representations of cultural knowledge. Using Sparse Autoencoders (SAEs), we find a set of features that encode different kinds of information pertaining to a culture. SAEs are well suited to this task because they decompose model activations into largely monosemantic, human-interpretable features, unlike individual transformer neurons, which are often highly polysemantic \cite{cunningham2023sparse, bricken2023monosemantic}. We aggregate these features into country-level \textbf{Cu}ltural \textbf{E}mbeddings (\method), which serve as compact representations of the cultural knowledge encoded within a model. An overview of our approach is shown in Figure~\ref{fig:main}.

We use \method in three ways that leverage the interpretability of SAE features. First, we analyze the cultural concepts encoded by the features themselves uncovering coherent cultural structure within the model’s representations. For example, geographically related cultures share
features capturing regional cuisine, geography, and ethnic references:
East Asian countries share features related to Asian ethnicity and
Chinese geographic locations; Mediterranean countries share
features referencing olives and olive oil production; most countries share a tea-related feature etc. We also identify features that are uniquely associated with individual countries, capturing more specific cultural concepts such as cuisine, institutions, and geopolitical topics. We also observe a clear hierarchy across layers: early layers encode lexical identity markers, while later layers capture richer cultural semantics.

Second, we use \method to diagnose implicit cultural defaults in model
generations. Under underspecified prompts, models overwhelmingly default
to Anglophone cultures: 33.1\% of responses align most closely with the U.S. and 26.9\% with the U.K., together accounting for roughly\textbf{ 60\%} of cultural defaults. Third, we construct white-box steering interventions that amplify culturally salient features during generation. Steering substantially improves cultural localization across models. On Gemma-2-9B, steered generations win \textbf{48\%} of cultural faithfulness comparisons against explicit prompting baselines (vs. 24\% losses) and \textbf{53\%} of rarity comparisons (vs. 17\% losses). These gains
persist even when steering is applied on top of explicit cultural
prompts, showing that feature-level interventions complement existing
black-box localization methods while providing fine-grained control
over cultural alignment.

%% file: sections/3-methodology.tex
\section{Methodology}
\label{sec:method}
An overview of our methodology is shown in Figure~\ref{fig:main}. We discover
latent features that capture culture-specific signals in LLM representations
and aggregate them into sparse, interpretable vectors called
\textbf{Cu}ltural \textbf{E}mbeddings (\method). These embeddings provide compact representations of cultural information and form the basis for both bias
analysis and steering toward a target culture.

\noindent \textbf{Data Source}: We use the CANDLE dataset \cite{nguyen2023extracting}, which contains cultural commonsense assertions across 240+ countries and regions. We retain only countries with at least 500 assertions to ensure sufficient coverage, resulting in the following set: \textit{Australia, Canada, China, Egypt, England, France,
Germany, Greece, India, Ireland, Israel, Italy, Japan, Mexico, Russia, Scotland,
South Korea, Spain, Thailand, Turkey, United Kingdom, United States}. The raw
data contains substantial redundancy and many trivial statements (e.g.,
\emph{``Canadian whisky is made in Canada''}), which provide limited cultural
signal. To address this, we uniformly sample 100 assertions per country to
ensure diversity. We further augment each country with 100 LLM-generated
assertions produced using GPT-4.1, designed to surface more culturally salient
and rarer concepts. The original 100 assertions are provided as in-context examples to avoid duplication. We also perform a post-verification of the cultural faithfulness of generated assertions using Gemini-2.5-Pro, giving us a final total count of 4,334 assertions across 22 countries. The prompt used and a detailed analysis of the dataset and features selected with and without the augmented data can be found in Appendix \ref{sec:augment}.

\subsection{Step 0: Can models classify assertions into cultures?}
\label{sec:step0}
Our goal of finding features that belong to certain cultures rests on the assumption that models encode culture-specific information in a way that permits discrimination across them.  To test this, we sample data from the assertion corpus described above and remove explicit country names to avoid trivial cues (Appendix \ref{sec:augment}). We then pose this as a classification task to the model -- given a rewritten cultural assertion, can the model predict its country of origin from a fixed set of country labels? Note that this is a multi-class classification problem with 22 labels (a random baseline would achieve 4.5 F1 points). To understand which layers most encode cultural information, we train linear probes on residual stream activations at different layers. This allows us to not only validate the presence of discriminative cultural signals in model representations but also provides insight into where it might be encoded.

\subsection{Step 1: Feature Activation Extraction}

Given the assertion dataset, we extract SAE feature activations across all layers (for which SAEs are available). For each input $x$, this yields a $d_{\text{sae}}$-dimensional activation vector $a(x)$ per layer, where each dimension records the activation of one feature.  These activations provide the raw signals from which \method embeddings are constructed. For each text input, we record feature activations at every token. Since salient features may appear at different positions, we apply \emph{max pooling} across the sequence, i.e., for feature $f_j$ and input $x$, we compute
\[
a_j(x) = \max_{t \in \text{tokens}(x)} z_j(t),
\]
where $z_j(t)$ is the activation of feature $f_j$ at token $t$. The resulting vector $a(x)$ represents the max-pooled activation of the assertion in the SAE feature space.


\subsection{Step 2: Constructing \method}
\label{sec:cue}
To determine which SAE features are most indicative of the culture label, we compute the mutual information (MI) between each feature’s activation and the country label:
\[
I(A_j; C) = \sum_{a_j, c} P(a_j, c) \log \frac{P(a_j, c)}{P(a_j) P(c)}.
\]
A high value of $I(A_j; C)$ indicates that feature $f_j$ provides significant information about the cultural label. We rank all features globally by MI and select features using a cumulative MI threshold $\rho$. Specifically, we include the highest-MI features until their
cumulative MI reaches a fraction $\rho$ of the total MI. Using a cumulative threshold $\rho$ allows the number of selected features to adapt to the overall MI distribution rather than requiring a fixed feature budget. We denote this index set by $\mathcal{S}$.

For each input $x$, we define its \method representation by restricting the activation vector to the selected feature set $\mathcal{S}$:
\[
\method(x) = a(x)[\mathcal{S}] \in \mathbb{R}^{|\mathcal{S}|}.
\]
Intuitively, $\method(x)$ captures how strongly each culturally informative
feature is activated for the input.

We construct a prototype for each country $c \in \mathcal{C}$ by averaging
\method representations over all associated assertions:
\[
\mathbf{p}_{\method}^{(c)} =
\frac{1}{N_c} \sum_{x \in \mathcal{D}_c} \method(x),
\]
where $\mathcal{D}_c$ is the set of assertions labeled with country $c$ and
$N_c = |\mathcal{D}_c|$. Each prototype is a sparse, interpretable vector
summarizing the typical activation pattern of culturally salient features for
that country. These prototypes form the basis for both bias analysis and
steering.

\subsection{Step 3a: \method for Bias Analysis}
Having constructed country-level prototypes, we first use \method to diagnose
implicit cultural biases exhibited by the model under underspecified prompts.
This step is purely analytical and does not involve any intervention.

SAE features may capture both culture-specific signals and general semantic
patterns that occur across all cultures (e.g., food or social interaction).
To isolate relative cultural signals and remove shared, culture-agnostic
activation patterns, we subtract a global baseline by computing the mean
prototype across all countries:
\[
\boldsymbol{\mu}_{\text{global}} =
\frac{1}{|\mathcal{C}|} \sum_{c \in \mathcal{C}} \mathbf{p}_{\method}^{(c)}.
\]

Centering representations in this way removes global activation biases that
are shared across classes, a common property of neural representation spaces
\cite{ethayarajh2019contextual}. We then center each country prototype:
\[
\tilde{\mathbf{p}}^{(c)} =
\mathbf{p}_{\method}^{(c)} - \boldsymbol{\mu}_{\text{global}}.
\]

Given a generated response $y$, we extract SAE activations and apply max
pooling as in Step~1 to obtain $a(y)$. Restricting to the salient features and
applying the same centering yields:
\[
\tilde{\mathbf{a}}(y) =
a(y)[\mathcal{S}] - \boldsymbol{\mu}_{\text{global}}.
\]

We quantify cultural alignment toward country $c$ using cosine similarity:
\[
\text{bias}(y,c) =
\cos\bigl(\tilde{\mathbf{a}}(y), \tilde{\mathbf{p}}^{(c)}\bigr).
\]

Higher similarity indicates that the activation pattern of the generated
response aligns more closely with the direction defined by that country's
prototype in SAE feature space. This allows us to identify which cultural
prototype the model implicitly defaults to under underspecified prompts.

\subsection{Step 3b: \method for Steering}
While Step~3a diagnoses cultural defaults, we now use \method to actively steer
generation toward a target culture through white-box interventions on internal
representations.

For a target culture $c_{\text{target}}$, we compute the mean prototype of all
non-target cultures:
\[
\bar{\mathbf{p}}_{\method}^{(\text{others})} =
\frac{1}{|\mathcal{C} \setminus \{c_{\text{target}}\}|}
\sum_{c \neq c_{\text{target}}}
\mathbf{p}_{\method}^{(c)}.
\]

We then define a feature-space steering direction by taking the difference
between the target prototype and this baseline:
\[
\boldsymbol{\delta} =
\mathbf{p}_{\method}^{(c_{\text{target}})} -
\bar{\mathbf{p}}_{\method}^{(\text{others})}.
\]

This vector highlights SAE features that are characteristic of the target
culture while suppressing those common to other cultures.

To apply steering within the model, we decode this feature-space direction into
the residual stream using the SAE decoder
$W_{\text{dec}} \in \mathbb{R}^{d_{\text{sae}} \times d_{\text{model}}}$:
\[
\mathbf{v}_{\method} =
W_{\text{dec}}^\top \boldsymbol{\delta}.
\]

At inference time, let $\mathbf{h}$ denote the current residual activation.
We apply steering via:
\[
\mathbf{h}' = \mathbf{h} + \alpha \mathbf{v}_{\method},
\]

where $\alpha$ controls the strength of the intervention and is selected
empirically. Varying $\alpha$ allows continuous control over the degree of
cultural localization in the generated output. 

We explain the methodology intuitively end-to-end using a hypothetical example in Appendix \ref{app:toy_example} for interested readers.

%% file: sections/4-experiments.tex
\section{Experimental Setup}
\label{sec:setup}
We evaluate our approach across multiple open-weight LLMs and model scales to assess the robustness of \method-based bias analysis and steering. All experiments are conducted using pretrained models and SAEs without any additional fine-tuning. 

\vspace{2mm}
\noindent\textbf{Models:} We run experiments on three model families -- \texttt{Gemma-2-2B}, \texttt{Gemma-2-9B}, and \texttt{LLaMA-3.1-8B} and use their pretrained SAEs from Hugging Face.\footnote{\url{https://huggingface.co/google/gemma-scope}; \url{https://huggingface.co/OpenMOSS-Team/Llama-Scope}} For \texttt{Gemma-2-2B}, we use pretrained SAEs with widths of 16K and 65K; for \texttt{Gemma-2-9B}, we use widths of 16K and 131K; and for \texttt{LLaMA-3.1-8B}, we use widths of 32K and 128K since pretrained SAEs are available for all layers for these widths. We refer to these variants as \textsc{G2-2B-16K}, \textsc{G2-2B-65K}, \textsc{G2-9B-16K}, \textsc{G2-9B-131K}, \textsc{L3.1-8B-32K}, and \textsc{L3.1-8B-128K}, respectively. We specifically select SAE configurations that are available for \emph{all layers} of the corresponding model. This design choice allows us to later analyze the layer-wise distribution of culturally informative features discovered by \method. For Gemma models, we use the instruction-tuned variants to support open-ended generation; SAEs trained on the base models have been shown to transfer effectively to these variants. Additional hyperparameter details can be found in Appendix \ref{app:hparam}.

\vspace{2mm}
\noindent\textbf{Baseline Settings:} We evaluate generation under two prompting conditions following the explicit versus implicit cue distinction in \citet{veselovsky2025localized}. In the \textsc{Implicit} setting, prompts contain no explicit country or cultural cues. In the \textsc{Explicit} setting, prompts include an explicit country identifier of the form ``I am from \texttt{<country>}''. Steering is applied in both settings by adding the steering vector to the residual stream, yielding four experimental conditions: \textsc{Implicit}, $\textsc{Steer}_{\textsc{Implicit}}$, \textsc{Explicit}, and $\textsc{Steer}_{\textsc{Explicit}}$. Our primary comparison examines how steering in the absence of cues ($\textsc{Steer}_{\textsc{Implicit}}$) compares with explicit prompting alone (\textsc{Explicit}), which serves as a natural prompt-augmentation baseline. Empirically, we also find that steering complements explicit prompting: even when cultural cues are provided, steering ($\textsc{Steer}_{\textsc{Explicit}}$) yields additional gains and enables fine-grained control over the degree of cultural alignment. We therefore report results across all four settings and demonstrate how steering provides a continuous control knob for cultural localization (Figure~\ref{fig:main}). Bias analysis is conducted across all four settings to characterize how prompting and steering independently and jointly influence implicit cultural alignment. 

\vspace{2mm}
\noindent\textbf{Evaluation Metrics:} We evaluate open-ended LLM generations on the creative writing prompts from \citet{veselovsky2025localized} (listed in Table~\ref{tab:creative_prompts}) along three dimensions: (i) \emph{cultural faithfulness}, which measures how well a generation reflects the intended target culture; (ii) \emph{rarity}, which captures whether generations surface less common or long-tail cultural concepts; and (iii) \emph{fluency}, which measures grammaticality and overall
coherence. We use fluency only as a filtering criterion when sweeping steering strengths $\alpha$. Steering strengths that consistently produce low fluency generations are discarded to avoid degenerate outputs. 

Because these prompts produce open-ended creative outputs for which reference answers are unavailable, we adopt the \emph{LLM-as-a-judge} evaluation paradigm.  To improve robustness and reduce reliance on a single evaluator, we use an ensemble of two independent LLM judges. Each judge scores generations on 1--10 Likert scale for cultural faithfulness, rarity, and fluency using the same evaluation rubric. The final scalar score for each dimension is computed as the mean of the two judges. Using multiple judges helps mitigate idiosyncratic biases that can arise in individual LLM evaluators and produces more stable evaluation signals.

\emph{Pairwise preference evaluation} -- In addition to scalar scoring, we perform pairwise preference comparisons. For each prompt, the judges are presented with two responses generated under different experimental conditions (e.g., \textsc{Explicit} vs.\ $\textsc{Steer}_{\textsc{Implicit}}$) and are asked to select which response better satisfies the evaluation criteria. Pairwise evaluation has been widely adopted in modern LLM benchmarks, including MT-Bench and Chatbot Arena, where a judge selects the preferred response between two candidates \cite{zheng2023judging}. Each pairwise comparison is evaluated independently by both judges. If the judges select the same response, that response is recorded as the winner; if the judges disagree, the comparison is recorded as a tie. We report win rates across prompts for each condition and treat these pairwise comparisons as the primary reliability check for the scalar evaluation results. To mitigate position bias in pairwise evaluation, the order of responses is randomized before being presented to the judges.

The exact evaluation prompts used for the judges are provided in Appendix \ref{app:eval_prompts}.

\begin{figure*}[t]
    \centering
    \includegraphics[width=0.98\linewidth]{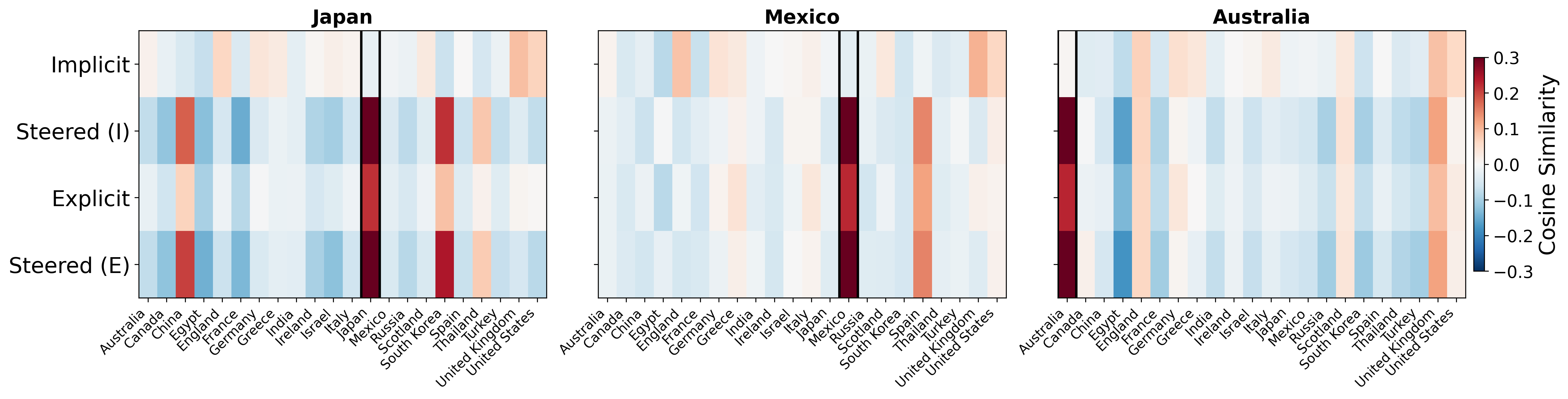}
    \caption{Heatmaps show cosine similarity between generated responses and country prototypes under \textsc{Implicit}, $\textsc{Steer}_{\textsc{Implicit}}$, \textsc{Explicit}, and $\textsc{Steer}_{\textsc{Explicit}}$ conditions. 
Implicit prompting concentrates alignment on Anglophone countries, while explicit prompting and steering progressively redistribute similarity mass toward target cultures and away from default cultural priors.}
    \label{fig:bias}
\end{figure*}

\begin{figure*}[t]
    \centering
    \includegraphics[width=0.98\linewidth]{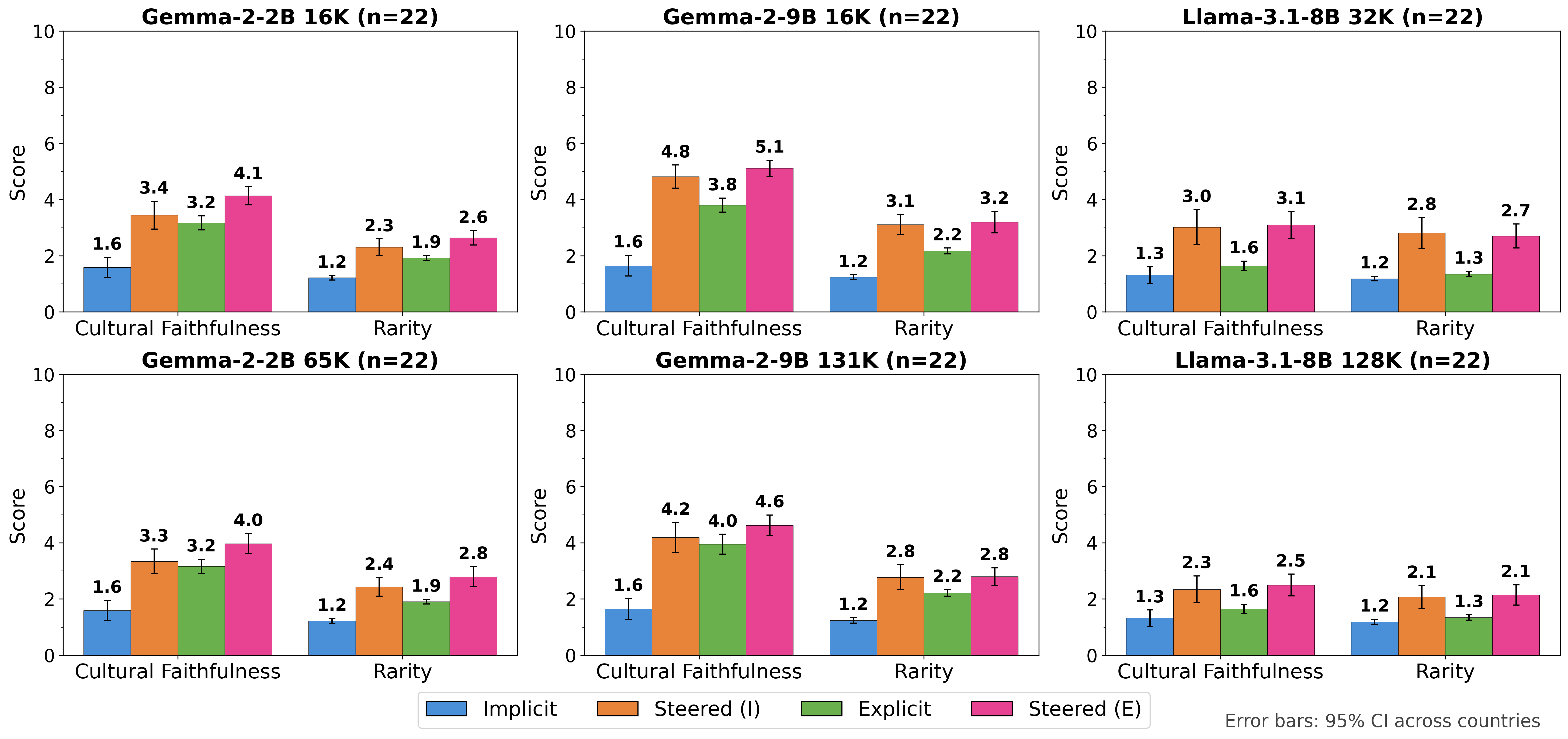}
    \caption{Mean cultural faithfulness and rarity scores (1--10 scale) across four settings---\textsc{Implicit}, $\textsc{Steer}_{\textsc{Implicit}}$, \textsc{Explicit}, and $\textsc{Steer}_{\textsc{Explicit}}$. Steered conditions consistently improve both metrics over their unsteered counterparts, with gains scaling with model capacity.}
    \label{fig:ind_score}
\end{figure*}

\begin{figure*}[t]
    \centering
    \includegraphics[width=0.98\linewidth]{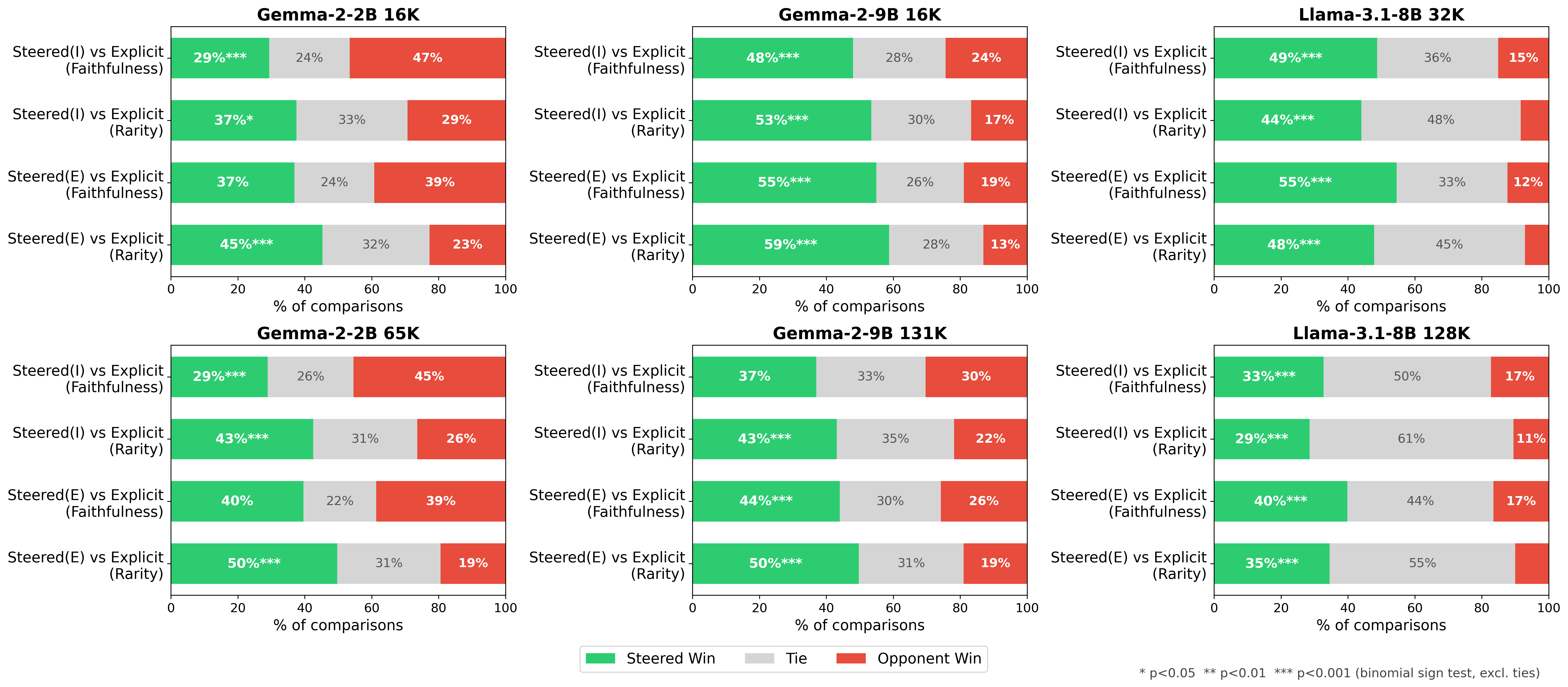}
    \caption{Pairwise win/tie/loss rates (\%) comparing steered outputs against explicit prompting with an ensemble of judges. Green bars indicate steered wins; red bars indicate explicit prompting wins. Steering outperforms explicit prompting on both faithfulness and rarity across most configurations, with the strongest gains on Gemma-2-9B 16K and Llama-3.1-8B 32K.}
    \label{fig:pairwise}
\end{figure*}

\section{Results and Analysis}
\label{sec:results}
A complete set of results are in Figures \ref{fig:ind_score}, \ref{fig:pairwise} and \ref{fig:bias}. First, we probe model representations to test whether culture-specific signals are encoded and where they emerge (RQ1). Next, we do a detailed analysis of high MI features and study their behaviour through Neuronpedia (RQ2),\footnote{\url{https://www.neuronpedia.org/gemma-scope}} examine cultural defaults in underspecified prompts and quantify these biases (RQ3). We then apply \textsc{CuE}-based interventions to evaluate whether steering can elicit rare, long-tail knowledge while remaining culturally faithful (RQ4) and analyse per-culture variations in RQ5.

\paragraph{RQ1: Do LLMs encode culture-specific signals, and where are they located?} Yes. A linear classifier trained to predict the country of origin of cultural assertions achieves \textbf{53.2\% macro-F1} on \texttt{Gemma-2-9B} (chance: \textbf{4.5\%} over 22 classes), indicating that culture-specific information is linearly decodable from model representations. Note that our goal here is not to achieve high F1 scores. In fact, this is practically impossible since there are many assertions with a single culture label that are in-fact \emph{cross-cultural} (e.g., “tea is popular”) and thus not uniquely identifying. Layer-wise probing shows that classification performance improves in later layers, but remains strong across the entire network, exceeding 30\% macro-F1 at all layers. Based on this, we apply steering across all layers rather than restricting interventions to only the top of the network. Full layer-wise results and confusion matrices are reported in Appendix \ref{app:cfn}.

\paragraph{RQ2: What concepts do shared vs.\ unique features encode?}

To interpret the cultural concepts encoded by SAE features selected via mutual information, we query Neuronpedia's automated explanations for features in the \texttt{gemma-scope-9b-pt-res-16k} dictionaries across all 42 residual stream layers. For each country, we identify features with the highest activation relative to the cross-country mean, weighted by mutual information score. We organize findings by whether features are \emph{shared} across culturally related countries or \emph{unique} to individual nations, and examine both positively and negatively associated features. Throughout, we reference individual SAE features using the notation \texttt{layer:index} (e.g., \texttt{22:6129} denotes feature 6129 in layer 22).

We find that shared features cluster along recognizable cultural--geographic boundaries. East Asian countries (China, Japan, South Korea, Thailand) share features encoding Asian ethnicity references (\texttt{22:6129}), Chinese geographical locations (\texttt{24:5526}), and Asian cultural elements (\texttt{30:10797})---notably with China-centric features dominating the cluster. British Isles nations (England, Scotland, Ireland, UK) share features for British culture (\texttt{27:9710}) and political entities (\texttt{37:3658}). Continental European countries (France, Germany, Italy, Spain) share features for European nationalities (\texttt{16:12833}) and multilingual identity (\texttt{12:2050}). We found an interesting feature common to Mediterranean countries (Greece, Italy, Spain, Turkey) encoding references to \emph{olives and olive oil production} (\texttt{38:10325})\footnote{\url{https://www.neuronpedia.org/gemma-2-9b/38-gemmascope-res-16k/10325}}, while a \emph{tea and tea-related concepts} feature (\texttt{33:5320}) activates broadly across 18 of 22 countries with strongest activation for England, China, and Turkey---suggesting the model has learned genuine cultural--agricultural groupings beyond surface-level country-name detection. 

Negatively associated features reveal complementary structure: British identity features such as \texttt{37:3658} and \texttt{36:14590} show strongly below-average activation in East Asian country contexts, while Chinese geography features (\texttt{40:1773}, \texttt{39:4317}) are similarly suppressed for Continental European and British contexts---indicating the model represents cultural identity partly through \emph{absence} of features associated with distant cultural groups. Country-unique features capture fine-grained cultural semantics: Indian cuisine and institutions (\texttt{35:7065}), Japanese governance (\texttt{39:2895}), Russian sanctions and geopolitics (\texttt{37:4570})\footnote{\url{https://www.neuronpedia.org/gemma-2-9b/37-gemmascope-res-16k/4570}}, and Middle Eastern cuisine (\texttt{35:14071}). We also observe a clear layer hierarchy: early layers (0--15) capture lexical identity markers (e.g., the token ``Indian,'' \texttt{3:6007}), while later layers (33--41) encode compositional cultural semantics such as cuisine, institutions, and geopolitical framing. A comprehensive feature catalog with Neuronpedia links is provided in Appendix~\ref{app:feature_catalog}.

\paragraph{RQ3: What cultural defaults do LLMs exhibit in underspecified prompts and how does steering/prompting change this?}

We analyse one model (Gemma-2-9B) here, but observe similar trends across models. Models default overwhelmingly to Anglophone cultures. Under implicit prompting, 33.1\% of responses align most closely with the US and 26.9\% with the UK, together accounting for 60\% of all cultural defaults. To quantify this skew, we compute a \emph{bias concentration index} that measures how far each country's share deviates from a perfectly even split across all 22 countries ($\approx$4.5\% each); a score of 0 means all cultures are equally represented, while higher scores indicate that a few cultures dominate. The implicit baseline scores 0.333, non-Western cultures such as Egypt, South Korea, and Turkey are weakly activated as defaults.

We also calculate this index for the rest of the experimental settings (taking responses across all countries into account). Explicit cultural prompting partially mitigates this bias: the concentration index drops to 0.115, though the US and UK still dominate at 19.8\% and 15.5\%, respectively. Steering provides substantially stronger debiasing---under $\textsc{Steer}_{\textsc{Implicit}}$, the concentration index falls to 0.051, an \textbf{84.7\%} reduction relative to the unsteered baseline. The most balanced condition is $\textsc{Steer}_{\textsc{Explicit}}$, which achieves a concentration index of 0.045, with a more even distribution across countries including Mexico (5.5\%), China (5.1\%), and Greece (6.4\%) alongside the US (13.1\%) and UK (10.7\%).

\paragraph{RQ4: Can \method steer model outputs toward target cultures?}

Yes. Steering strength $\alpha$ is auto-selected per country from $\{0.25, 0.5, 1.0, 2.0\}$ by choosing the smallest $\alpha$ whose steered implicit cultural score exceeds the explicit baseline while maintaining fluency $\geq 5.0$; the selected values range from 0.25 (8 countries, e.g., China, Japan, UK) to 1.0 (Russia), with the majority at 0.5 (13 countries). Pairwise comparisons  (Figure~\ref{fig:pairwise}) demonstrate that steering consistently improves cultural localization beyond what prompting alone achieves, especially for larger models. 

For the best configuration (Gemma-2-9B, 16K SAE), $\textsc{Steer}_{\textsc{Implicit}}$ wins \textbf{48\%} of faithfulness comparisons against explicit prompting versus only 24\% for explicit (28\% ties), and \textbf{53\%} of rarity comparisons versus 17\% for explicit ($p < 0.001$). The effect is even stronger when combining steering with explicit prompts: $\textsc{Steer}_{\textsc{Explicit}}$ wins \textbf{55\%} of faithfulness and \textbf{59\%} of rarity comparisons against explicit alone, confirming that steering is complementary to prompting.

Notably, rarity improvement is more robust than faithfulness across all configurations---even Gemma-2-2B achieves net positive rarity gains (37\% win vs 29\% loss)---suggesting that steering reliably surfaces more distinctive cultural content even when overall faithfulness gains are limited by model capacity.

\paragraph{RQ5: How do steering effects vary across cultures?}

Steering effectiveness varies substantially across countries (Appendix \ref{app:results}). We analyze per-country win/tie/loss breakdowns for the best-performing configuration (Gemma-2-9B, 16K SAE) and identify three patterns:

\textbf{a) High-benefit cultures} -- Countries including Thailand (19 wins, 3 ties, 3 losses out of 25), Australia (17/6/2), Greece (16/6/3), South Korea (15/7/3), Mexico (16/4/5), and India (15/5/5) show strong faithfulness improvement even against the explicit prompting baseline. Rarity gains are more pronounced, with several countries showing near-universal wins (e.g., Thailand 22/2/1, India 21/2/2, South Korea 18/6/1). These countries have low baseline cultural scores (1.0--1.8/10), suggesting the model possesses relevant cultural knowledge but does not activate it by default; steering successfully elicits this latent knowledge. Supporting this, steering toward these countries increases similarity to culturally proximate neighbors (e.g., steering toward Japan raises alignment with South Korea and China, as shown in Figure~\ref{fig:bias}), indicating that steering amplifies coherent cultural subspaces.

\textbf{b) Already-represented cultures} -- The United States (10/9/6 on faithfulness) shows modest gains with many ties, reflecting the well-documented Anglophone default bias (RQ3)---the model already produces US-centric content, yielding ties when steered content matches the default. Rarity is even more tie-dominated (5/14/6), confirming that the US baseline already produces distinctive content. Canada (9/10/6) and England (12/9/4) show similar patterns with substantial tie rates.

\textbf{c) Benefit only with explicit prompting} -- Egypt and Israel show a different pattern. With $\textsc{Steer}_{\textsc{Implicit}}$ alone, Egypt \emph{loses} to explicit prompting (5/9/11 on faithfulness), and Israel is similarly negative (5/10/10). However, $\textsc{Steer}_{\textsc{Explicit}}$---which combines steering with an explicit country mention reverses this: Egypt improves to 13/7/5 and Israel to 9/10/6. Russia follows a similar pattern (10/5/10 $\rightarrow$ 11/6/8). This suggests that for these cultures, the steering vector encodes relevant cultural knowledge but is insufficient to \emph{identify} the target culture on its own---the model needs the explicit cue to ``anchor'' the cultural direction, after which steering amplifies it. Why this happens remains unclear, but it would be worth exploring in future work.

Rarity improvement is the most universal effect: $\textsc{Steer}_{\textsc{Explicit}}$ achieves net positive rarity gains (wins $>$ losses) for all 22 countries. Even for challenging cases like Russia (11/8/6) and Israel (9/12/4), steered responses surface more distinctive cultural content than explicit prompting alone.

%% file: sections/2-related_work.tex
\section{Related Work}

\paragraph{Cultural bias and alignment in LLMs:} LLMs often default to Western or Anglo-centric norms across languages and tasks, reflecting imbalances in training corpora. Recent surveys emphasize that most evaluations operationalize ``culture'' via proxies and black-box probes, leaving model internals largely unexamined \citep{adilazuarda2024towards, liu2025culturally}. Prompt- and persona-based localization can reduce bias but frequently amplifies stereotypes or oversimplifies national identities \citep{cheng2023marked, gupta2023bias}. Post-training alignment similarly risks collapsing diverse cultural preferences into a single averaged objective \citep{alkhamissi2024cultural}. Dataset efforts such as CANDLE expose systematic gaps in cultural commonsense knowledge, which in turn propagate into model generations \citep{nguyen2023extracting,bhatia2023gd}. These limitations motivate methods that go beyond prompting or fine-tuning to diagnose and control cultural behaviour at the representational level.

\paragraph{Interpretability and sparse autoencoders:} Mechanistic interpretability seeks to uncover how models implement behaviors rather than only observing outputs. Probes, attention maps, and neuron analyses provide correlational insights but struggle with polysemanticity, where single units encode many unrelated concepts \citep{cunningham2023sparse}. Sparse autoencoders (SAEs) address this by decomposing hidden states into sparse, overcomplete dictionaries whose features are more monosemantic and human-interpretable \citep{bricken2023monosemantic}. Benchmarks and scaling studies demonstrate that SAEs extract coherent features across large LLMs and allow direct causal interventions, while also identifying challenges in evaluation and feature stability \citep{karvonen2025saebench,shi2025route}. Compared to attribution or linear probes, SAEs provide more faithful, manipulable latent directions for steering model behavior.

\paragraph{Cultural NLP x Interpretability:} Work at this intersection remains nascent. Recent studies such as \textit{Culturescope} reveal how underrepresented cultural knowledge becomes entangled with dominant representations via patching analyses, but they remain diagnostic rather than interventional \citep{zhang2025culturescope}. The closest work to ours is that of \citep{veselovsky2025localized}, who also obtain contrastive cultural activation vectors by prompting the model with and without country names, and use that to intervene on model generations. However, our work is the first to leverage SAE-derived cultural features both to diagnose implicit cultural defaults and to steer outputs. We are also able to discover interpretable features across multiple SAE layers, which helps verify the correctness of our methodology. By directly modulating interpretable latent features, we bridge interpretability and alignment, offering a white-box approach to culturally faithful generation.

%% file: sections/6-conclusion.tex
\section{Conclusion}
We introduced Cultural Embeddings (\method), an interpretability-driven
framework for discovering and manipulating culturally relevant features
in large language models. By leveraging sparse autoencoders, we identify
interpretable latent features that encode culturally salient concepts
and use them to study how cultural knowledge is represented within LLMs.

Our analysis reveals that cultural information is structured in the
model’s internal representations. Shared features capture regional
patterns such as cuisine, geography, and ethnicity across related
cultures, while country-specific features encode finer-grained concepts
such as institutions, governance, and geopolitical topics. These
features also exhibit a layer-wise hierarchy, where early layers capture
lexical identity markers and later layers encode richer cultural
semantics.

Using these interpretable features, we diagnose strong cultural defaults
in model generation, where underspecified prompts disproportionately
align with Anglophone cultural prototypes. Feature-level interventions
allow us to modify these internal representations and improve cultural
localization, consistently surfacing rarer cultural concepts and often
improving cultural faithfulness.

More broadly, our results suggest that cultural knowledge in LLMs is
partially latent but structurally organized. Interpretable feature
analysis provides a direct way to uncover this structure and offers a
promising direction for building more transparent and controllable
language models. Future work could extend this approach to richer
cultural definitions beyond countries, explore how cultural features
emerge during training, and use interpretability tools to better audit
and align global AI systems.

%% file: sections/7-ethics_limitations.tex
\section{Limitations}

\paragraph{Culture as country:} We use \emph{country} as a proxy for culture, following much prior work in cultural NLP \citep{adilazuarda2024towards,cheng2023marked,mukherjee2023global}. This choice was pragmatic given available datasets, but it inevitably flattens intra-country diversity and risks essentializing identity \citep{zhou2025culture}. Despite this, we would like to note that our method itself is generalizable and flexible by design for any label set (e.g., regions, communities, linguistic groups), provided we have data for these labels to find features using.

\paragraph{Hyperparameter sensitivity:} Steering requires tuning the strength parameter $\alpha$ and selecting suitable SAE layers. While this can be a large search space to optimize for, this is still not as challenging as prompt engineering or alignment methods, where the “right” phrasing or reward remains uncertain \citep{ouyang2022training}. Importantly, our search space is finite and interpretable, unlike the vast and unstructured space of natural language prompts.


\section{Ethical Considerations}

\paragraph{Dual use:} Cultural steering can enhance inclusivity by making LLMs more contextually appropriate, but it also risks reinforcing essentialist or stereotypical portrayals if misapplied \citep{blodgett2020language,alkhamissi2024cultural}. Over-steering could produce caricatures or amplify cultural tropes, and optimizing for one group may disadvantage others.

\paragraph{Safeguards:} Responsible use requires participatory design with cultural stakeholders \citep{mohammad2021ethics}, transparency about how culture is operationalized \citep{gebru2021datasheets}, and ongoing monitoring for harm. Documentation of cultural datasets and steering parameters is especially important to avoid misuse.

\paragraph{Interpretability:} Our SAE-based approach provides white-box levers for cultural alignment, but interpretability can also be abused. Adversaries could manipulate features to generate misinformation or offensive outputs. Further, even though SAE features are monosemantic to a higher degree than transformer neurons, they still hold risks of polysemanticity and we may unintentionally affect other behaviours.